\RequirePackage{fix-cm}
\documentclass[twocolumn]{svjour3}    

\usepackage{amsmath}
\let\vec\overrightarrow 
\usepackage{rotating,tabularx}
\usepackage{epsfig}
\usepackage{mathptmx}
\usepackage{graphicx}
\usepackage{amsmath}
\usepackage{amssymb}
\usepackage{times}
\usepackage{epsfig}
\usepackage{graphicx}
\usepackage{algorithm}
\usepackage{algorithmic}
\usepackage{array}
\usepackage{multirow}
\usepackage[utf8]{inputenc}
\usepackage[T1]{fontenc}
\usepackage{lmodern} 
\usepackage{pbox}
\usepackage{fixltx2e}
\usepackage{mathtools}
\usepackage{color}
\usepackage{hyperref}
\usepackage{fancyhdr}
\usepackage[font=footnotesize, caption=false]{subfig}
\usepackage{url}
\usepackage{epstopdf}
\usepackage{pdflscape}
\usepackage{tablefootnote}
\usepackage{blindtext}
\usepackage[output-decimal-marker={.}]{siunitx}

\DeclareMathOperator*{\argmin}{argmin}

\usepackage{microtype} 
\usepackage{lipsum}

\journalname{Machine Vision and Applications} 

\begin{document}

\title{Solving the Robot-World Hand-Eye(s) Calibration Problem with Iterative Methods}

\author{Amy Tabb \and Khalil M. Ahmad Yousef} 

\institute{Amy Tabb  \at
	United States Department of Agriculture\\
	Agricultural Research Service\\
	Appalachian Fruit Research Laboratory\\
	Kearneysville, West Virginia, 25430, USA.\\
	\email{amy.tabb@usda.gov}\\
	Mention of trade names or commercial products in this publication is solely for the purpose of
	providing specific information and does not imply recommendation or
	endorsement by the U.S. Department of Agriculture.  USDA is an equal
	opportunity provider and employer.           \\
	A. Tabb acknowledges the support of United States National Science Foundation grant number IOS-1339211.  
	\and
	Khalil M. Ahmad Yousef \at
	Computer Engineering Department\\
	The Hashemite University\\
	Zarqa 13115, Jordan.\\
	\email{khalil@hu.edu.jo}\\
	The citation information for this paper is:\\
 A. Tabb and K. M. Ahmad Yousef, “Solving the robot-world hand-eye(s) calibration problem with iterative methods,” Machine Vision and Applications, vol. 28, no. 5, pp. 569–590, Aug. 2017. \href{http://doi.org/10.1007/s00138-017-0841-7}{doi: 10.1007/s00138-017-0841-7}\\
We include an Erratum to this version of the paper.  Please see Section \ref{changelog} for details.
}

\date{Received: 28 July 2016 / Accepted: 1 April 2017}

\maketitle

\begin{abstract}
Robot-world, hand-eye calibration is the problem of determining the transformation between the robot end-effector and a camera, as well as the transformation between the robot base and the world coordinate system.  This relationship has been modeled as $\mathbf{AX}=\mathbf{ZB}$, where $\mathbf{X}$ and $\mathbf{Z}$ are unknown homogeneous transformation matrices.  The successful execution of many robot manipulation tasks depends on determining these matrices accurately, and we are particularly interested in the use of calibration for use in vision tasks.  In this work, we describe a collection of methods consisting of two cost function classes, three different parameterizations of rotation components, and separable versus simultaneous formulations.  We explore the behavior of this collection of methods on real datasets and simulated datasets, and compare to seven other state-of-the-art methods.  Our collection of methods return greater accuracy on many metrics as compared to the state-of-the-art.  The collection of methods is extended to the problem of robot-world hand-multiple-eye calibration, and results are shown with two and three cameras mounted on the same robot.
\end{abstract}

\section{Introduction} \label{sec:Intro}
\begin{figure}
\centering
\includegraphics [scale=0.8]{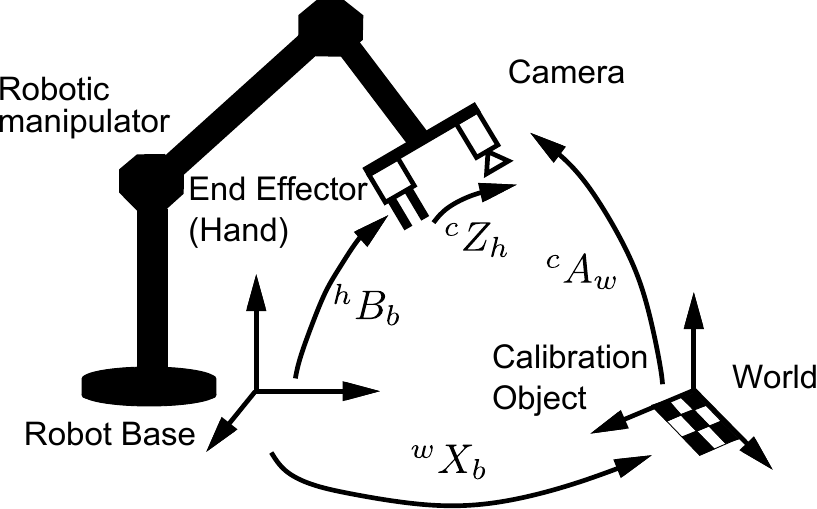}
\caption{{Hand-Eye Robot-World Calibration: A camera (eye) mounted at the
end-effector (hand) of a robot.}}
\label{fig:transformation_relationships}
\end{figure}
The robot-world hand-eye calibration problem  consists of determining the homogeneous transformation matrices (HTMs) between the robot hand, or end-effector, to the camera, as well as the transformation of the robot base to the world coordinate system.   

The preliminaries of the robot-world hand-eye calibration problem are as follows.  Let the transformation from the hand coordinate frame to the camera coordinate frame be $^{c}Z_{h}$ or simply $\mathbf{Z}$, and the transformation from the robot-base coordinate frame to the world coordinate frame be $^{w}X_{b}$ or simply $\mathbf{X}$. The transformation of the robot base frame to the hand coordinate frame is $^{h}B_{b}$ or simply $\mathbf{B}$, and is assumed to be known from the robot controller.  Finally, the transformation from the world coordinate frame to the camera coordinate frame is represented by
HTM $^{c}A_{w}$ or $\mathbf{A}$.  $\mathbf{A}$ is calculated using a camera calibration procedure such as Zhang \cite{zhang2000flexible}, where the world coordinate frame is defined by a calibration object in the workspace.  The transformations are illustrated by Figure \ref{fig:transformation_relationships}.  We note here that the labeling of the transformations in our version of the problem is different than that used in the traditional robot-world hand-eye calibration problem \cite{Zhuang1994simultaneous}
in that some matrices are inverted ($\mathbf{A}$, $\mathbf{B}$) and  
the rest of matrices are exchanged ($\mathbf{X}$, $\mathbf{Z}$).  Despite this difference, the linear relationship $\mathbf{AX} = \mathbf{ZB}$ is still the same in our interpretation of the problem versus others.  We interpreted the problem differently because by doing so we are able to simplify the derivation of some of the cost functions that we use in Section \ref{ss:ReprojectionError}.

Given these preliminaries, Zhuang \textit{et al.} \cite{Zhuang1994simultaneous} formalized the robot-world hand-eye calibration explicitly as the homogeneous matrix equation $\mathbf{AX}=\mathbf{ZB}$, where all of the matrices are $4 \times 4$ matrices, and were the first to provide a method to find solutions for $\mathbf{X}$ and $\mathbf{Z}$.  Many different positions of the robot and camera are used to generate multiple relationships $\mathbf{A}_i\mathbf{X}=\mathbf{Z}\mathbf{B}_i$, $i \in [0, n-1]$, where $n$ is the number of robot poses used for the calibration.

The robot-world hand-eye calibration problem is different, though related, to the hand-eye calibration problem, which was formulated as $\mathbf{AX}=\mathbf{XB}$ by Shiu and Ahmad \cite{shiu1989calibration}.  In the hand-eye calibration problem, the matrices $\mathbf{A}_i$ and $\mathbf{B}_i$ are now considered as the relative transformations from one pose to another.  Use of relative transformations is problematic as decisions must be made as to how to convert absolute transformations into relative ones. This work considers instead $\mathbf{A}_i$ and $\mathbf{B}_i$ as poses with respect to the world coordinate system and robot base coordinate system, respectively, and as a result robot-world hand-eye calibration is needed.

We structure the paper as follows. In the rest of Section \ref{sec:Intro} and specifically in Subsection \ref{section:recent_work}, we provide an overview of existing approaches and recent work. In Subsection \ref{section:contributions}, we discuss our approach and contributions to the robot-world hand-eye calibration problem. Section \ref{section:rotation_param} introduces two classes of proposed methods and extends those classes to the multiple camera case.  Section \ref{section:Experiments-ErrorMetrics} describes the experiments, metrics, and implementation details. In Section \ref{section:Res}, the results on real and simulated datasets are shown and discussed. Finally, in Section \ref{sec:conclusions}, we present our conclusions.  

\subsection{Recent work}
\label{section:recent_work}

Most recent work on the robot-world hand-eye calibration problem makes use of the decomposition of $\mathbf{AX}=\mathbf{ZB}$ into a purely rotational part and a translational part, where $\mathbf{R}_A$ represents a $3 \times 3$ rotation matrix, and $\mathbf{t}_A$ a $3 \times 1$ translation vector as shown in Equation \ref{eq:decoupling_into_parts}. 
\begin{align}
\begin{bmatrix*}[l] \mathbf{R}_A &  \mathbf{t}_A \\ 0^T & 1 \end{bmatrix*}\begin{bmatrix*}[l] \mathbf{R}_X &  \mathbf{t}_X \\ \mathbf{0}^T & 1 \end{bmatrix*}&= \begin{bmatrix*}[l] \mathbf{R}_Z &  \mathbf{t}_Z \\ \mathbf{0}^T & 1 \end{bmatrix*} \begin{bmatrix*}[l] \mathbf{R}_B &  \mathbf{t}_B \\ \mathbf{0}^T & 1 \end{bmatrix*} \label{eq:decoupling_into_parts}
\end{align}

Methods utilizing the decomposition of the problem that is shown in Equation \ref{eq:decoupling_into_parts} into two parts are called separable methods. Separable methods cause Equation \ref{eq:decoupling_into_parts} to produce two other equations: the rotation part as represented by Equation \ref{eq:rotation_parts} and the translation part as represented by Equation \ref{eq:translation_parts}.  
\begin{align}
\mathbf{R}_A \mathbf{R}_X&=\mathbf{R}_Z \mathbf{R}_B \label{eq:rotation_parts}\\
\mathbf{R}_A \mathbf{t}_X + \mathbf{t}_A &=  \mathbf{R}_Z \mathbf{t}_B + \mathbf{t}_Z \label{eq:translation_parts}
\end{align}

Since Equation \ref{eq:translation_parts} is linear in $\mathbf{t}_X$ and $\mathbf{t}_Z$ if $\mathbf{R}_Z$ is known, the most frequent approach to estimate $\mathbf{X}$ and $\mathbf{Z}$ is to first find $\mathbf{R}_X$ and $\mathbf{R}_Z$  using Equation \ref{eq:rotation_parts}, and then use that solution and Equation \ref{eq:translation_parts} to find $\mathbf{t}_X$ and $\mathbf{t}_Z$.

Of those methods that separate the estimation of the rotation and translation parts using Equation \ref{eq:decoupling_into_parts}, there are several different approaches.  In Zhuang \textit{et al.} \cite{Zhuang1994simultaneous}, a linear solution was proposed for finding the unknowns $\mathbf{R}_X$ and $\mathbf{R}_Z$ by representing the rotation matrices as quaternions, and the translation components were found using linear least squares. Dornaika and Horaud in \cite{dornaika1998simultaneous} gave a closed-form method for estimating the rotation components using quaternions, which did not require normalization like in \cite{Zhuang1994simultaneous}. Translation components were estimated with linear least squares as in \cite{Zhuang1994simultaneous}.  Hirsh \textit{et al.} in \cite{hirsh2001iterative} proposed a separable, iterative approach in which the estimation of $\mathbf{X}$ and $\mathbf{Z}$ is alternated. The method consists of assuming that $\mathbf{Z}$ is known and estimating $\mathbf{X}$ by averaging $\mathbf{X}_i = \mathbf{Z}\mathbf{B}_i\mathbf{A}^{-1}$ for all $i$ to generate an estimate for $\mathbf{X}$; this process is repeated in a similar way for $\mathbf{Z}$ using the estimate found for $\mathbf{X}$, and the estimation of $\mathbf{X}$ and $\mathbf{Z}$ is continued until termination conditions are met.  Like the other methods in this group, estimation of rotation is separated from translation, and rotation is represented by quaternions.  In the method of Shah \cite{Shah2013Solving}, the Kronecker product and singular value decomposition were used to create a closed-form solution.

The second class of methods are typically called simultaneous solutions; they do not decouple rotation and translation error and thus have the advantage that error from the rotation estimation is not propagated to the translation estimation.  In \cite{dornaika1998simultaneous}, in addition to a closed-form separable solution, Dornaika and Horaud present a formulation of the problem as a non-linear least squares problem.  The rotation components are represented by matrices, so there are 18 parameters for rotation and 6 for translation.  The cost function contains penalty terms that enforce orthonormality of the rotation matrices.  Following Dornaika and Horaud, Stobl and Hirzinger in \cite{Strobl2006Optimal} estimate the translation and rotation components through non-linear estimation techniques, though in their approach weights for the rotational and translational parts are chosen automatically and a position/orientation precision ratio parameter is required. The work of Li \textit{et al.} \cite{li2010simultaneous} proposed a simultaneous solution that is found using dual quaternions and the Kronecker product.

Besides those approaches that deal with the equality $\mathbf{AX}= \mathbf{ZB}$ in separable and simultaneous versions, there have been some approaches within the context of the hand-eye calibration problem to refine the estimation of the camera calibration parameters. For example, in \cite{Horaud1995Hand}, Horaud and Dornaika incorporated camera calibration parameters into a cost function for the hand-eye calibration problem $\mathbf{AX} = \mathbf{XB}$ and solved using the Levenberg-Marquardt method when the rotations are represented by quaternions. In this method, the requirement that the quaternion must have unit norm was enforced using penalty terms. Recently, Malti \cite{malti2013hand} in a work about hand-eye calibration, incorporated a variation of the robot-world hand-eye calibration formulation and used reprojection error together with epipolar constraints to simultaneously refine the camera intrinsic and distortion parameters in addition to the $\mathbf{X}$ and $\mathbf{Z}$ HTMs.  Unlike Horaud and Dornaika \cite{Horaud1995Hand}, Malti in \cite{malti2013hand} uses Euler angles to represent rotation and as a result his method does not need penalty terms.

\subsection{Our approach and contributions}
\label{section:contributions}

Our work on the robot-world hand-eye calibration problem is motivated by three particular situations.  The first is that we use a robot-mounted camera for multi-view high-quality reconstruction of complicated objects such as leafless trees as in Tabb \cite{Tabb2013};  while some extrinsic camera calibration errors can be tolerated, the reconstruction outcomes improve when the camera calibration error is low.  The second situation is where we use the same robot-mounted camera in laboratory or field conditions, where it can be the case that non-experts will be involved with performing calibration and acquiring data from a remote location.  For this particular situation, our goal was to devise methods for robot-world hand-eye calibration that are not sensitive to the particular sequence of motions used during the calibration and is robust to non-ideal calibration situations. Finally third, we have some situations in which multiple cameras are mounted on the same robot end-effector.  While it is possible to only calibrate one camera using robot-world hand-eye calibration and then perform stereo camera calibration, we instead desired a way to calibrate all components at the same time so that error from one step is not propagated to the final calibration.  This approach also allows cameras to have vastly different fields of view and does not require overlapping fields of view for any two cameras, which is more flexible than stereo calibration.

We note here that the robot-world hand-eye calibration problem has been discussed in the literature for more than two decades.  Early contributions focused on linear and/or closed-form solutions because of computational efficiency.  However, with the advent of open source nonlinear least square solvers, such as {\textit{levmar}} \cite{lourakis04LM} and Ceres \cite{ceressolver}, in addition to the general interest in bundle adjustment \cite{Triggs1999Bundle}, iterative methods with previously unconsidered parameterizations of the rotational components offer advantages over the linear and closed-form approaches.

In this paper, we propose a collection of iterative methods for the robot-world hand-eye calibration problem. There are two classes of cost functions in this collection.  The first class of cost functions minimizes the sum of squared difference between $\mathbf{AX}$ and $\mathbf{ZB}$ over $n$ positions of the robot as given in Equation \ref{eq:general_cost_function} below, or in a closely-related version of it, where $F$ denotes the Frobenius norm. This class is discussed in Subsection \ref{class1}.
\begin{equation}
\sum_{i \in [0, n-1]} ||\mathbf{A}_i\mathbf{X}-\mathbf{Z}\mathbf{B}_i||^2_{{F}}
\label{eq:general_cost_function}
\end{equation}

However, in the first class of cost functions as given in Equation \ref{eq:general_cost_function}, we observed that artifacts and errors related to the calibration object and camera calibration method are sometimes propagated to the estimation of $\mathbf{X}$ and $\mathbf{Z}$.  The second class of cost functions aims to reduce the influence of these camera calibration artifacts by finding $\mathbf{X}$ and $\mathbf{Z}$ based on camera reprojection error, without explicitly using $\mathbf{A}$ and this is discussed in Subsection \ref{ss:ReprojectionError}.

Each of the aforementioned two classes of cost functions contains two or more sub-classes or methods, first, to explore different choices of cost functions; separable versus simultaneous solutions, and second, to explore different 
parameterizations of the rotation components. For the latter, we use three different possible parameterizations namely: Euler angles, axis-angle, and quaternions.
 
Since the two classes of cost functions are nonlinear, we use nonlinear least-square solvers based on the {Levenberg-Marquardt} method to find the approximate solutions of $\mathbf{X}$ and $\mathbf{Z}$.   
It is important to mention here that the above collection of classes and methods easily extend to the problem of calibrating multiple cameras mounted on one robot as shall be discussed in Subsection \ref{ss:MultCameras}.

Similarities between the state-of-the-art and our proposed collection of methods are as follows.  Our methods are iterative, like the method of Hirsh \textit{et al.} \cite{hirsh2001iterative} and the iterative method of Dornaika and Horaud \cite{dornaika1998simultaneous}. The work of Malti \cite{malti2013hand} concerning using camera reprojection error to estimate HTMs $\mathbf{X}$ and $\mathbf{Z}$ is similar to our second class of methods,  when using the Euler angle representation for rotations (Subsection \ref{ss:ReprojectionError}).  One difference is that Malti's work assumes that the motions are relative and as a result uses different definitions of the matrices than we do.

Our contributions to the state-of-the-art in robot-world hand-eye calibration are summarized as follows:
\begin{itemize}

\item{We provide a comprehensive comparison that explores different choices of cost function, parameter choices, and separable versus simultaneous solutions within our collection of methods, and contrast those methods with the state-of-the-art methods, on real datasets as well as on simulated datasets.}
\item{Our collection of methods is provided to the community as open source code \cite{tabb2017solving_dataset}.}
\end{itemize}

A portion of the work presented in this paper was previously published as \cite{Tabb2015Parameterizations}.  However, the present paper expands the work of \cite{Tabb2015Parameterizations} in the following ways:
\begin{itemize}
\item{This paper evaluates three different parameterizations of the rotation matrices in the experimental results: Euler angles, the axis-angle representation, and quaternions. Also, it discusses the effect of the parameter choice on results.  In contrast, the prior work only used Euler angle parameterizations. }
\item{While in \cite{Tabb2015Parameterizations} only simultaneous methods were proposed, we also created separable formulations of some of the cost functions and as a result were able to demonstrate the differences on the results produced by separability.}
\item{We incorporated a comparison with the work of Shah \cite{Shah2013Solving}, which was not presented in \cite{Tabb2015Parameterizations}.}
\item{Prior work only considered one camera. In contrast, the methods presented in this paper are generalized to the multiple-eye, one robot problem, with no limit to the number of cameras used.}
\item{Four additional real datasets were acquired.}
\item{Simulated datasets were added to the experiments, which allow comparisons between the estimated and true solutions and additional analysis of the methods.}
\end{itemize}

\section{Collection of methods description}
\label{section:rotation_param}

Many of the differences among the previous methods concern how orthonormality of the rotation matrices is maintained. In this section, we first describe the parameterizations of the rotational components of $\mathbf{X}$ and $\mathbf{Z}$ such that orthonormality is preserved, and then discuss in details the two proposed classes of cost functions that we briefly introduced in the previous section followed by the extension to multiple cameras.  

We explore three different parameterizations: Euler angle, axis-angle, and quaternions.  Since these rotations are commonplace in the robotics literature and practice, in the next paragraph, we will only give a brief sketch of some details relevant to such parameterizations. 

The Euler angle rotation representation is composed of three variables, which represent the rotation angles or direction cosines along the $x,y,z$ directions. In this work, we selected the directional order of the rotations in the following order: $x$, $y$, $z$.  Consequently, a rotation matrix can be represented as in the following manner $\mathbf{R} = R_Z R_Y R_X$.  The axis-angle representation is another three-variable representation, where a three-element vector $\mathbf{v}$ represents the axis of rotation, and $||\mathbf{v}||$ is the angle of rotation about the $\mathbf{v}$ axis such that a corresponding rotation matrix is generated with Rodrigues' rotation formula.  Finally, quaternions are a 4-element representation, where a unit vector $\mathbf{q}$ can be converted to an orthonormal matrix.

We now define some notation to help us describe our two classes of cost functions and their extension to multiple cameras.  Let the selected rotation representation be represented by a vector $\mathbf{p}$, with the size of $\mathbf{p}$ equal to $3$ for the Euler angles and the axis-angle representations, and equal to $4$ for the quaternion representation. Then the directed cosine matrix representation of $\mathbf{p}$ is $\mathbf{R}(\mathbf{p})$.

With these preliminaries, we will now present our first class of cost functions.

\subsection{First class: $\mathbf{AX}=\mathbf{ZB}$}
\label{class1}

The first class of cost functions uses the world to camera transformations $\mathbf{A}_i$ and base to end-effector transformations $\mathbf{B}_i$ to estimate the HTMs $\mathbf{X}$ and $\mathbf{Z}$. This is achieved by seeking the minimum of $c_1$ as given below:  
\begin{equation}
c_1 = \sum_{i = 0}^{n - 1} || \mathbf{A}_i\mathbf{X} - \mathbf{Z}\mathbf{B}_i||^2_{{F}}
\label{eq:c1_basic}
\end{equation}

In the cost function $c_1$ in Equation \ref{eq:c1_basic}, if we substitute-in the rotation representation $\mathbf{R}(\mathbf{p})$ and the translation vectors for the unknowns $\mathbf{X}$ and $\mathbf{Z}$, the minimization problem becomes:
\begin{equation}
\resizebox{\linewidth}{!}{ $
 \argmin\limits_{\mathbf{p}_X, \mathbf{t}_X, \mathbf{p}_Z, \mathbf{t}_Z} \underset{i=0}{\overset{n-1}{\sum}} \left\Vert \mathbf{A}_{i}\left[\begin{array}{cc}
\mathbf{R}(\mathbf{p}_X) & \mathbf{t}_{X}\\
\mathbf{0}^T & 1
\end{array}\right]-\left[\begin{array}{cc}
\mathbf{R}(\mathbf{p}_Z) & \mathbf{t}_{Z}\\
\mathbf{0}^T & 1
\end{array}\right]\mathbf{B}_{i}\right\Vert^2_{{F}}
$}
\label{eq:c1_param_simultaneous}
\end{equation}

A solution to the problem in Equation \ref{eq:c1_param_simultaneous} belongs to the simultaneous category of methods for robot-world hand-eye calibration, because the rotation and translation are solved for at the same time.

We can also break the minimization problem in Equation \ref{eq:c1_param_simultaneous} into two parts via a separable formulation: one part for rotation and the other part to estimate the translation components as captured by Equation \ref{eq:c1_param_separable_R} and Equation \ref {eq:c1_param_separable_t}, respectively.
\begin{equation}
\argmin\limits_{\mathbf{p}_X, \mathbf{p}_Z}\underset{i=0}{\overset{n-1}{\sum}} \left\Vert \mathbf{R}_{A, i}\mathbf{R}(\mathbf{p}_X) - \mathbf{R}(\mathbf{p}_Z) \mathbf{R}_{B, i} \right\Vert^2_{{F}}
\label{eq:c1_param_separable_R}
\end{equation}
\begin{equation}
\argmin\limits_{\mathbf{t}_X, \mathbf{t}_Z}\underset{i=0}{\overset{n-1}{\sum}} \left\Vert \mathbf{R}_{A, i}\mathbf{t}_X + \mathbf{t}_A -\mathbf{R}(\mathbf{p}_Z)\mathbf{t}_{B, i} - \mathbf{t}_Z \right\Vert^2_{{F}}
\label{eq:c1_param_separable_t}
\end{equation}

An approximate solution to Equation \ref{eq:c1_param_separable_R} is first found and then the translation components are found from Equation \ref{eq:c1_param_separable_t}.

The left hand $(\mathbf{AX})$ and right hand $(\mathbf{ZB})$ sides of the cost function $c_1$ represent the transformation from robot base to camera via the world coordinate system (left hand side) and from robot base to camera via the end-effector coordinate system (right hand side) as graphically captured in Figure \ref{fig:transformation_relationships}.  We also explore the use of a slightly different cost function $c_2$ as given in Equation \ref{eq:EulerC2}:
\begin{equation}
c_2 = \sum_{i = 0}^{n - 1} || \mathbf{A}_i - \mathbf{Z}\mathbf{B}_i\mathbf{X}^{-1}||^2_{{F}}
\label{eq:EulerC2}
\end{equation}

We can find an approximate solution to $c_2$ simultaneously, similar to what we did with the function $c_1$, by estimating $\mathbf{p}_X$, $\mathbf{t}_X$, $\mathbf{p}_Z$, and $\mathbf{t}_Z$.  In order to simplify the notation, we let $\tilde{\mathbf{X}} = \mathbf{X}^{-1}$.  Then $\tilde{\mathbf{p}}_X$ and $\tilde{\mathbf{t}}_X$ are the rotation parameters and translation vectors of $\mathbf{X}^{-1}$, respectively.  With these substitutions, we have the following minimization problem:

\begin{equation}
\resizebox{\linewidth}{!}{ $
 \argmin\limits_{\tilde{\mathbf{p}}_X, \tilde{\mathbf{t}}_X, \mathbf{p}_Z, \mathbf{t}_Z} \underset{i=0}{\overset{n-1}{\sum}} \left\Vert \mathbf{A}_{i} - \left[\begin{array}{cc}
\mathbf{R}(\mathbf{p}_Z) & \mathbf{t}_{Z}\\
\mathbf{0}^T & 1
\end{array}\right]\mathbf{B}_{i}\left[\begin{array}{cc}
\mathbf{R}(\tilde{\mathbf{p}}_X) & \tilde{\mathbf{t}}_{X}\\
\mathbf{0}^T & 1
\end{array}\right]\right\Vert^2_{{F}}
$}
\label{eq:c2_param_simultaneous}
\end{equation}

As with $c_1$, it is possible to create a separable formulation by first estimating the rotation components in Equation \ref{eq:c2_param_separable_R} and then the translation components in Equation \ref{eq:c2_param_separable_t}.

\begin{align}
\argmin\limits_{\tilde{\mathbf{p}}_X, \mathbf{p}_Z}& \underset{i=0}{\overset{n-1}{\sum}} \left\Vert \mathbf{R}_{A, i} - \mathbf{R}(\mathbf{p}_Z) \mathbf{R}_{B, i} \mathbf{R}(\tilde{\mathbf{p}}_X) \right\Vert^2_{{F}}
\label{eq:c2_param_separable_R}\\
\argmin\limits_{\tilde{\mathbf{t}}_X, \mathbf{t}_Z}&\underset{i=0}{\overset{n-1}{\sum}} \left\Vert \mathbf{t}_{A, i} -  \mathbf{R}(\mathbf{p}_Z) \mathbf{R}_{B, i} \tilde{\mathbf{t}}_X - \mathbf{R}(\mathbf{p}_Z) \mathbf{t}_B - \mathbf{t}_Z \right\Vert^2_{{F}}
\label{eq:c2_param_separable_t}
\end{align}

In both of the minimization formulations (simultaneous and separable) for $c_1$ and $c_2$, the number of parameters to be estimated for $\mathbf{X}$ and $\mathbf{Z}$ are: six translation components and six or eight rotation parameters, depending on the choice of the rotation representation.  Approximate solutions to all of the problems proposed in this paper are found with the {Levenberg-Marquardt} method for non-linear least squares \cite{marquardt1963algorithm} and specifically its implementation in \textit{Ceres} \cite{ceressolver}. For an initial solution, values are chosen such that $\mathbf{R}(\mathbf{p}_X)$ and $\mathbf{R}(\mathbf{p}_Z)$ are $3 \times 3$ identity matrices and $\mathbf{t}_X$ and $\mathbf{t}_Z$ have all elements zero.

\subsection{Second class: camera reprojection error} \label{ss:ReprojectionError}

In this subsection, we present our second class of cost functions and methods, which are based on the idea of minimizing camera reprojection error in order to find $\mathbf{X}$ and $\mathbf{Z}$. The approach presented in this subsection shares many similarities with the camera calibration approach of Zhang in \cite{zhang2000flexible}, where the extrinsic camera calibration parameters are estimated by minimizing the camera reprojection error.  Unlike the first class of methods, this approach is sensitive to the choice of initial solution; briefly, we use the result from the first class of methods as an initial solution and a more in-depth discussion of initial solution choices can be found in Subsection \ref{ss:implementation}.

Before getting into the details of this class of methods, we first mention some preliminaries and notation. Let a three-dimensional point on the calibration object be $\vec{\mathcal{X}}$, and because there are multiple $m$ such points we give a subscript $j \in [0, m-1]$,  $\vec{\mathcal{X}}_j$.  It is assumed that all of the points are detected in the images used for calibration, so in the case of the chessboard pattern, these points are the corners of the chessboard.   When $\vec{\mathcal{X}}_j$ is projected to the image from robot position $i$ using the intrinsic camera calibration parameters, we have image point $\vec{\tilde{\mathbf{x}}}_{ij}$ and a corresponding original image point $\vec{\mathbf{x}}_{ij}$.  We can represent $\vec{\tilde{\mathbf{x}}}_{ij}$ in the context of the robot-world hand-eye calibration problem as follows:
\begin{equation}
\vec{\tilde{\mathbf{x}}}_{ij} = f({\mathbf{k}}, \left[\mathbf{Z}\mathbf{B}_i\mathbf{X}^{-1}\right]_{3\times 4} \vec{\mathcal{X}}_j)
\label{eq:reprojection_with_k1}
\end{equation}
where ${\mathbf{k}}$ is a vector containing the intrinsic camera calibration parameters. The bracket notation $\left[\,\right]_{3\times 4}$ is used to denote the upper $3\times 4$ sub-matrix of what is inside the bracket. $f()$ is the function that transforms $\left[\mathbf{Z}\mathbf{B}_i\mathbf{X}^{-1}\right]_{3\times 4} \vec{\mathcal{X}}_j$ into image points using ${\mathbf{k}}$.

Concerning ${\mathbf{k}}$, we use $4$ parameters from the intrinsic camera calibration matrix and $8$ radial and tangential distortion parameters, so $|{\mathbf{k}}| = 12$.  However, other radial and tangential distortion models may be chosen with no change to the method. As with $c_2$ in Equation \ref{eq:c2_param_simultaneous}, to simplify the representation of Equation \ref{eq:reprojection_with_k1}, we substitute for $\mathbf{X}^{-1}$ another matrix $\tilde{\mathbf{X}}$, which is composed of an orthonormal rotation matrix and translation vector.

Given the preliminaries discussed above, the reprojection sum of squares error {(\textit{rsse})} is: 
\begin{equation}
rsse=\sum_{i = 0}^{n-1}\sum_{j=0}^{m-1}||\vec{\mathbf{x}}_{ij}-\vec{\tilde{\mathbf{x}}}_{ij}||^{2}
\label{eq:rsse}
\end{equation}
\noindent and by substituting $\vec{\tilde{\mathbf{x}}}_{ij}$ we have:
\begin{equation}
 rsse= \underset{i=0}{\overset{n-1}{\sum}}\underset{j=0}{\overset{m-1}{\sum}}||\vec{{\mathbf{x}}}_{ij} - 
f({\mathbf{k}}, \left[\mathbf{Z}\mathbf{B}_i\tilde{\mathbf{X}}\right]_{3\times 4} \vec{\mathcal{X}}_j)||^2
\label{eq:rsse_rwhec0}
\end{equation}
\noindent We note that in Equations \ref{eq:rsse} and \ref{eq:rsse_rwhec0} and for the remainder of this paper, we use the L2 norm for vectors.

Finally, if we substitute in the rotation representation $\mathbf{R}(\mathbf{p})$ and the translation vectors for the unknown $\tilde{\mathbf{X}}$  and $\mathbf{Z}$ matrices, the camera reprojection sum of squares error ($rsse$) minimization problem becomes as given in Equation \ref{eq:rsse_rwhec}, which we refer to as $rp_1$ method:

\begin{multline}
 \argmin\limits_{\tilde{\mathbf{p}}_X, \tilde{\mathbf{t}}_X, \mathbf{p}_Z, \mathbf{t}_Z} \underset{i=0}{\overset{n-1}{\sum}}\underset{j=0}{\overset{m-1}{\sum}}||\vec{{\mathbf{x}}}_{ij} - \\
f(\mathbf{k}, \left[\begin{array}{cc}
\mathbf{R}(\mathbf{p}_Z) & \mathbf{t}_{Z}\\
\mathbf{0}^T & 1
\end{array}\right]\mathbf{B}_{i}\left[\begin{array}{cc}
\mathbf{R}(\tilde{\mathbf{p}}_X) & \tilde{\mathbf{t}}_{X}\\
\mathbf{0}^T & 1
\end{array}\right]_{3\times 4} \vec{\mathcal{X}}_j||^2
\label{eq:rsse_rwhec}
\end{multline}
\noindent where $\tilde{\mathbf{p}}_X$ and $\tilde{\mathbf{t}}_X$ are again the rotation parameter vector and translation vectors of $\tilde{\mathbf{X}}$ or $\mathbf{X}^{-1}$, respectively.  

Within the simultaneous formulation of the robot-world hand-eye calibration problem as given in Equation \ref{eq:rsse_rwhec}, it is also possible to refine the camera intrinsic parameter vector ${\mathbf{k}}$ by letting ${\mathbf{k}}$ be parameters to be estimated in order to produce a local minimum in Equation \ref{eq:rsse_rwhec} instead of constants as given in Equation \ref{eq:rsse_rwhec_with_k}, which we refer to as the $rp_2$ method.
\begin{multline}
 \argmin\limits_{\tilde{\mathbf{p}}_X, \tilde{\mathbf{t}}_X, \mathbf{p}_Z, \mathbf{t}_Z, {\mathbf{k}}} \quad \underset{i=0}{\overset{n-1}{\sum}}\underset{j=0}{\overset{m-1}{\sum}}||\vec{{\mathbf{x}}}_{ij} - \\
f({\mathbf{k}}, \left[\begin{array}{cc}
\mathbf{R}(\mathbf{p}_Z) & \mathbf{t}_{Z}\\
\mathbf{0}^T & 1
\end{array}\right]\mathbf{B}_{i}\left[\begin{array}{cc}
\mathbf{R}(\tilde{\mathbf{p}}_X) & \tilde{\mathbf{t}}_{X}\\
\mathbf{0}^T & 1
\end{array}\right]_{3\times 4} \vec{\mathcal{X}}_j||^2
\label{eq:rsse_rwhec_with_k}
\end{multline}

In contrast with the minimized reprojection sum of squared error $(rsse)$ in both $rp_1$ and $rp_2$ methods, it is important to mention that in the camera calibration literature, the reprojection root mean squared error ($rrmse$) is more common, as shown in Equation \ref{eq:rmse}. The $rrmse$ represents the average Euclidean distance between detected and reprojected calibration pattern points in the image plane, and its units are pixels. However, since the parameters that result in a minimum of $rsse$ also result in a minimum of $rrmse$, $rsse$ is used as a cost function in this work. 
\begin{equation}
rrmse= \sqrt{\frac{1}{mn}\sum_{i = 0}^{n-1}\sum_{j=0}^{m-1}||\vec{\mathbf{x}}_{ij}-\vec{\tilde{\mathbf{x}}}_{ij}||^{2}}
\label{eq:rmse}
\end{equation}

\subsection{Extending the two proposed classes of methods to multiple cameras} \label{ss:MultCameras}

The multiple camera case requires estimating one HTM $\mathbf{X}$ and multiple HTMs $\mathbf{Z}$.  As such, if we let the number of cameras be $q$, and $\mathbf{A}_{i, 0}$ be the transformation from the world coordinate frame to the $0^{\text{th}}$ camera at the $i^{\text{th}}$ robot position.  Then the relationship between the matrices in the context of robot-world hand-multiple-eyes calibration problem would be formulated as: 
\begin{align}
\mathbf{A}_{i, 0}\mathbf{X} =& \mathbf{Z}_0\mathbf{B}\\
\mathbf{A}_{i, 1}\mathbf{X} =& \mathbf{Z}_1\mathbf{B}\\
\vdots& \nonumber\\
\mathbf{A}_{i, q-1}\mathbf{X} =& \mathbf{Z}_{q-1}\mathbf{B}
\label{eq:c1_basic_multiple}
\end{align}
for all $i \in [0, n-1]$.  Therefore, we should now be able to represent all of the cost functions presented thus far in Subsections \ref{class1} and \ref{ss:ReprojectionError} within this multiple camera context.  For instance, Equation  \ref{eq:c1_param_simultaneous} in the multiple camera context becomes: 
\begin{multline}
 \argmin\limits_{\mathbf{p}_X, \mathbf{t}_X, \mathbf{p}_{Z, 0}, \mathbf{t}_{Z, 0}, \mathbf{p}_{Z, 1}, \mathbf{t}_{Z, 1}, \dots, \mathbf{p}_{Z, q-1}, \mathbf{t}_{Z, q-1}}\\
 \underset{d=0}{\overset{q-1}{\sum}} \underset{i=0}{\overset{n-1}{\sum}} \left\Vert \mathbf{A}_{i, d}\left[\begin{array}{cc}
\mathbf{R}(\mathbf{p}_X) & \mathbf{t}_{X}\\
\mathbf{0}^T & 1
\end{array}\right]-\left[\begin{array}{cc}
\mathbf{R}(\mathbf{p}_{Z, d}) & \mathbf{t}_{Z, d}\\
\mathbf{0}^T & 1
\end{array}\right]\mathbf{B}_{i}\right\Vert^2_{{F}}
\label{eq:c1_param_simultaneous_multiple}
\end{multline}
\noindent where $\mathbf{p}_{Z, d}$ and $\mathbf{t}_{Z, d}$ are the rotation parameterization and translation vectors for the $d^{\text{th}}$ HTM, $\mathbf{Z}_{d}$.  Separable versions follow using the same pattern, as it is done in the cost function $c_1$ and the $rsse$ formulations of the two classes of methods found in the previous subsections.

When using multiple cameras, it is common that in order to gain a wide variety of views of the calibration object, not all images for each camera will view the calibration object, particularly if the cameras are far apart or have different lenses and imaging sensor sizes. We now describe a generic method that we used in our implementation, to weight the influences from each camera in the cost function.  We desired that each camera have an equal influence, but this weighting can be adjusted according to other needs. 

We let the set of robot positions where a camera $d$ can view the calibration object be $\mathbb{S}_d$, and denote $min_s$ as the minimum size of $\mathbb{S}_d$ determined for all cameras. Then, an individual weight $w_d$ for camera $d$ can be set as $\frac{min_s}{|\mathbb{S}_d|}$.  Furthermore, if we let the cost function that is to be minimized be $g()$, the formulation of the robot-world hand-multiple-eyes calibration problem such that missing observations are handled is given by Equation \ref{eq:c1_param_simultaneous_multiple_weights}.
\begin{equation}
 \argmin\limits_{\substack{\mathbf{p}_X, \mathbf{t}_X,\\ \mathbf{p}_{Z, 0}, \mathbf{t}_{Z, 0},\\ \mathbf{p}_{Z, 1}, \mathbf{t}_{Z, 1},\\ \dots\\ \mathbf{p}_{Z, q-1}, \mathbf{t}_{Z, q-1}}}\\
 \underset{d=0}{\overset{q-1}{\sum}} \underset{i \in \mathbb{S}_d}{{\sum}} w_d \hspace{0.1 cm} g(\mathbf{p}_X, \mathbf{t}_X, \mathbf{p}_{Z, d}, \mathbf{t}_{Z, d})
\label{eq:c1_param_simultaneous_multiple_weights}
\end{equation}

The problem in Equation \ref{eq:c1_param_simultaneous_multiple_weights} can be applied to the first class of cost functions, as well as the $rp1$ method in the second class of methods.  For the $rp2$ method in the second class of cost functions, we also estimate intrinsic camera calibration parameters $\mathbf{k}$ for all $q$ cameras and doing so follows from Equation \ref{eq:c1_param_simultaneous_multiple_weights}.


\section{Performance Evaluation on real datasets} \label{section:Experiments-ErrorMetrics}

\begin{figure}[H]
	\centering
	\subfloat[The arrangement of cameras in dataset 7 ]{\includegraphics[width=0.85\linewidth]{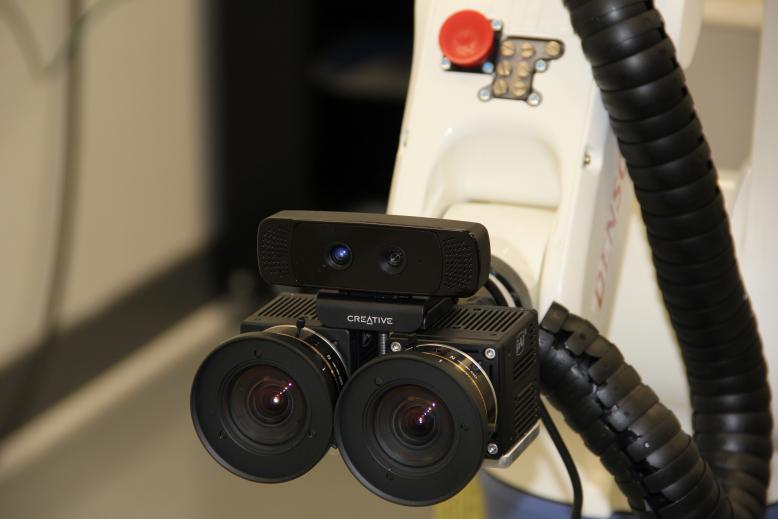}\label{sf:three_cameras}}
	
	\subfloat[Camera 0]{\includegraphics[width=0.30\linewidth]{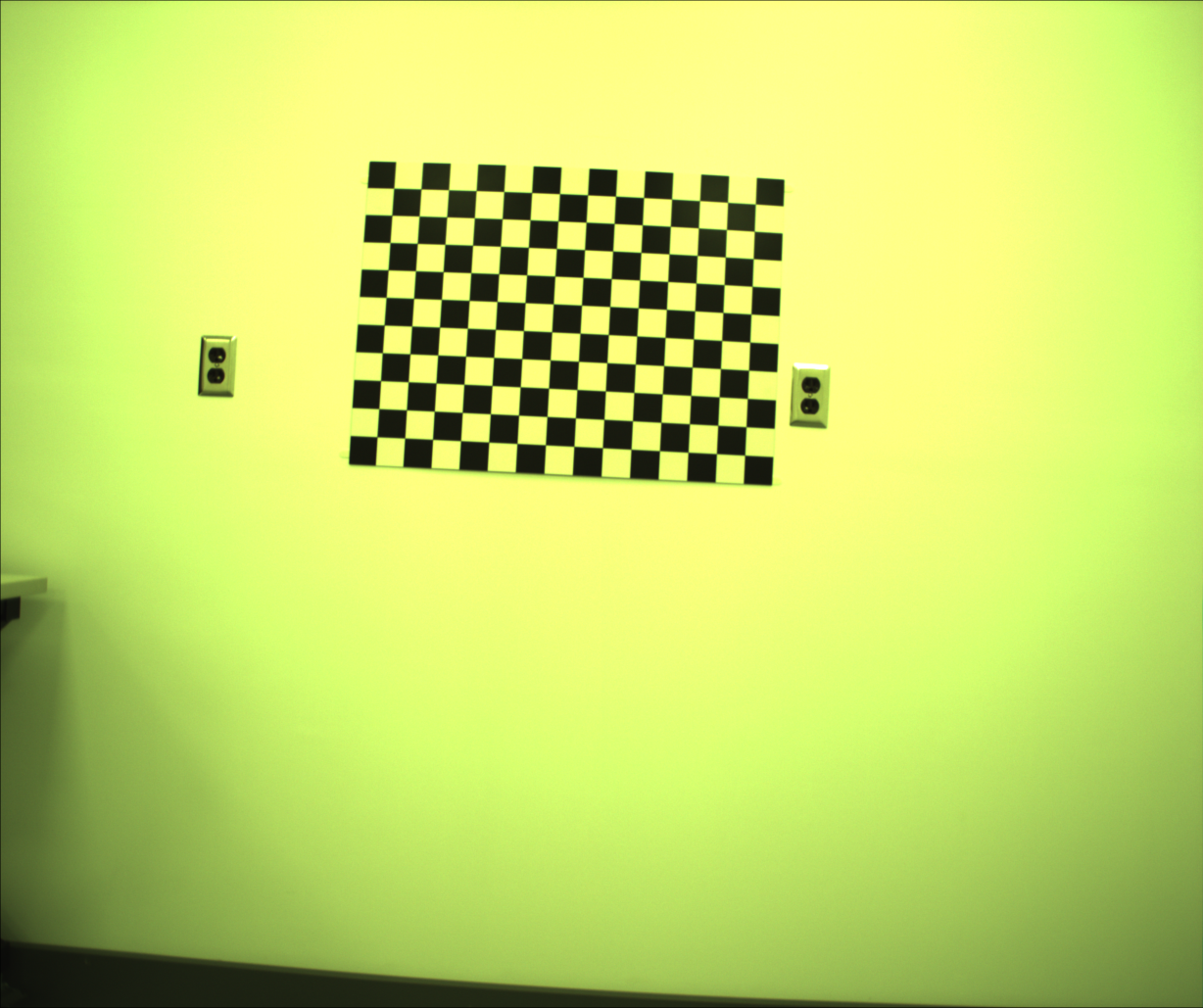}} \hspace{0.20 cm}
	\subfloat[Camera 1]{\includegraphics[width=0.30\linewidth]{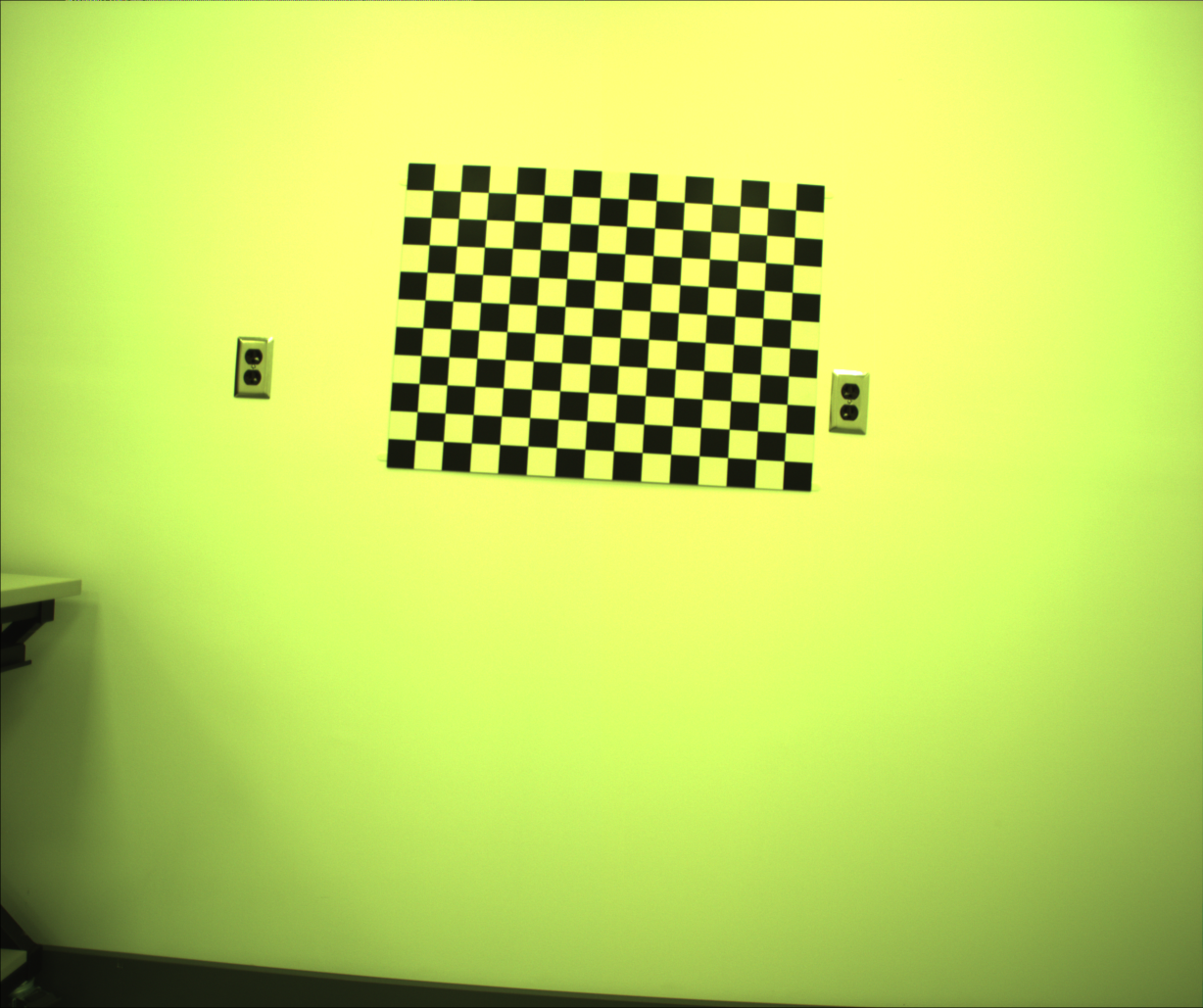}} \hspace{0.20 cm}
	\subfloat[Camera 2]{\includegraphics[width=0.33\linewidth]{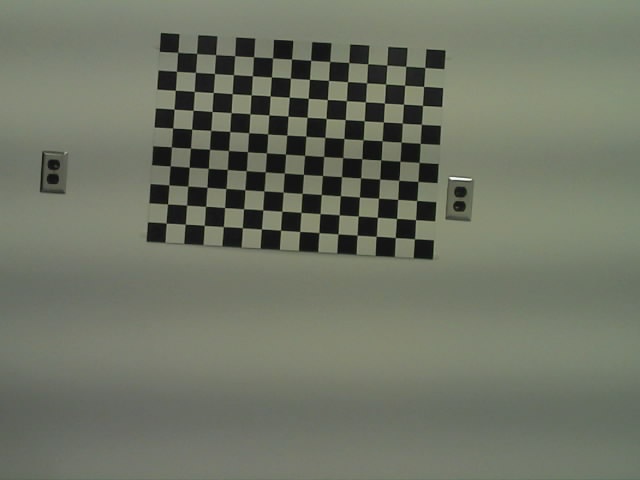}}
	\caption{An example of the experimental setup for dataset 7, consisting of three cameras and the images from each camera for one stop of the robot} 

	\label{fig:example_images}
\end{figure}

\begin{figure}
\centering
\subfloat[Denso VS-6577GM-B]{\includegraphics[width=0.85\linewidth]{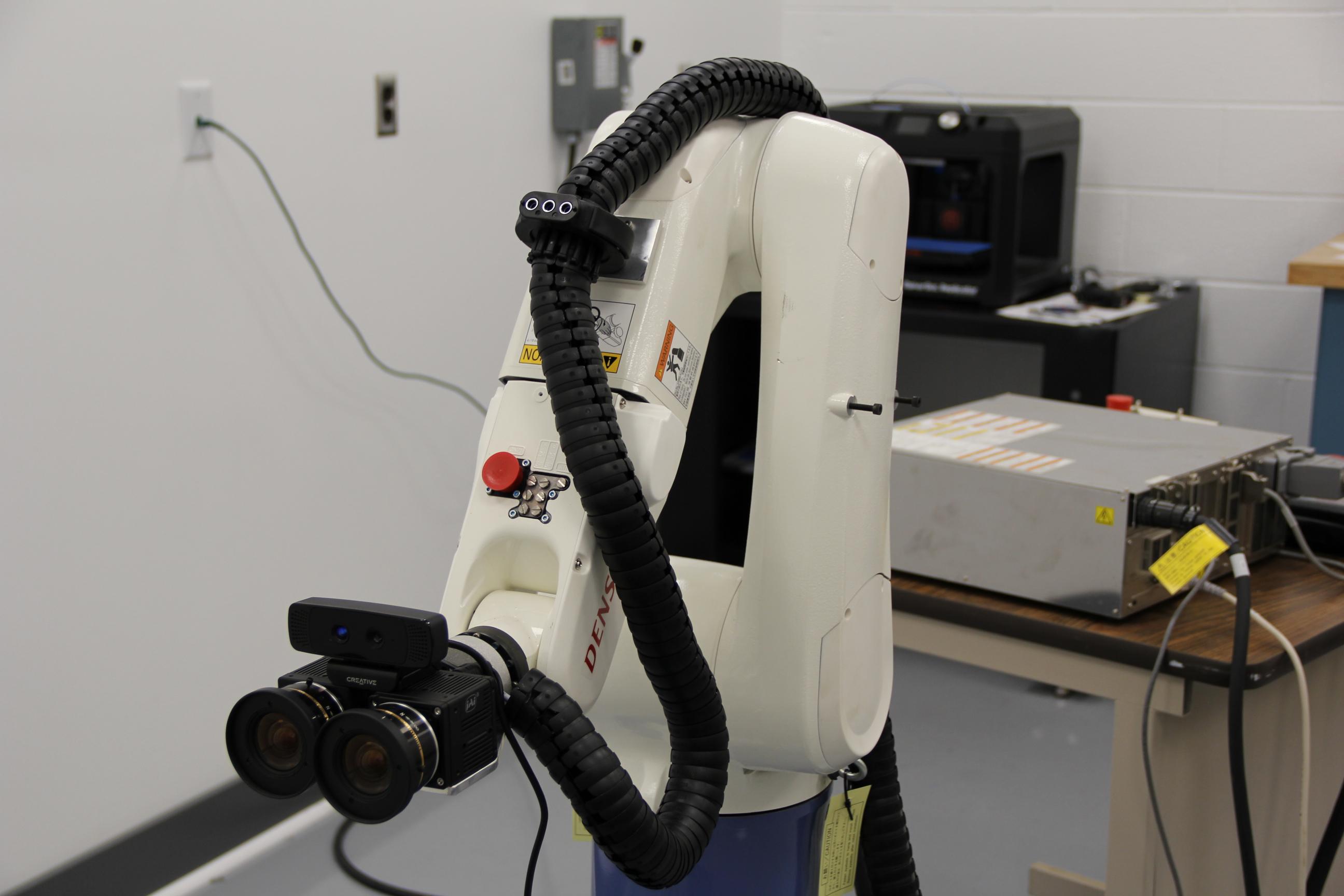}
\label{sf:small_robot}}

\subfloat[DENSO VM-60BIG]{\includegraphics[width=0.85\linewidth]{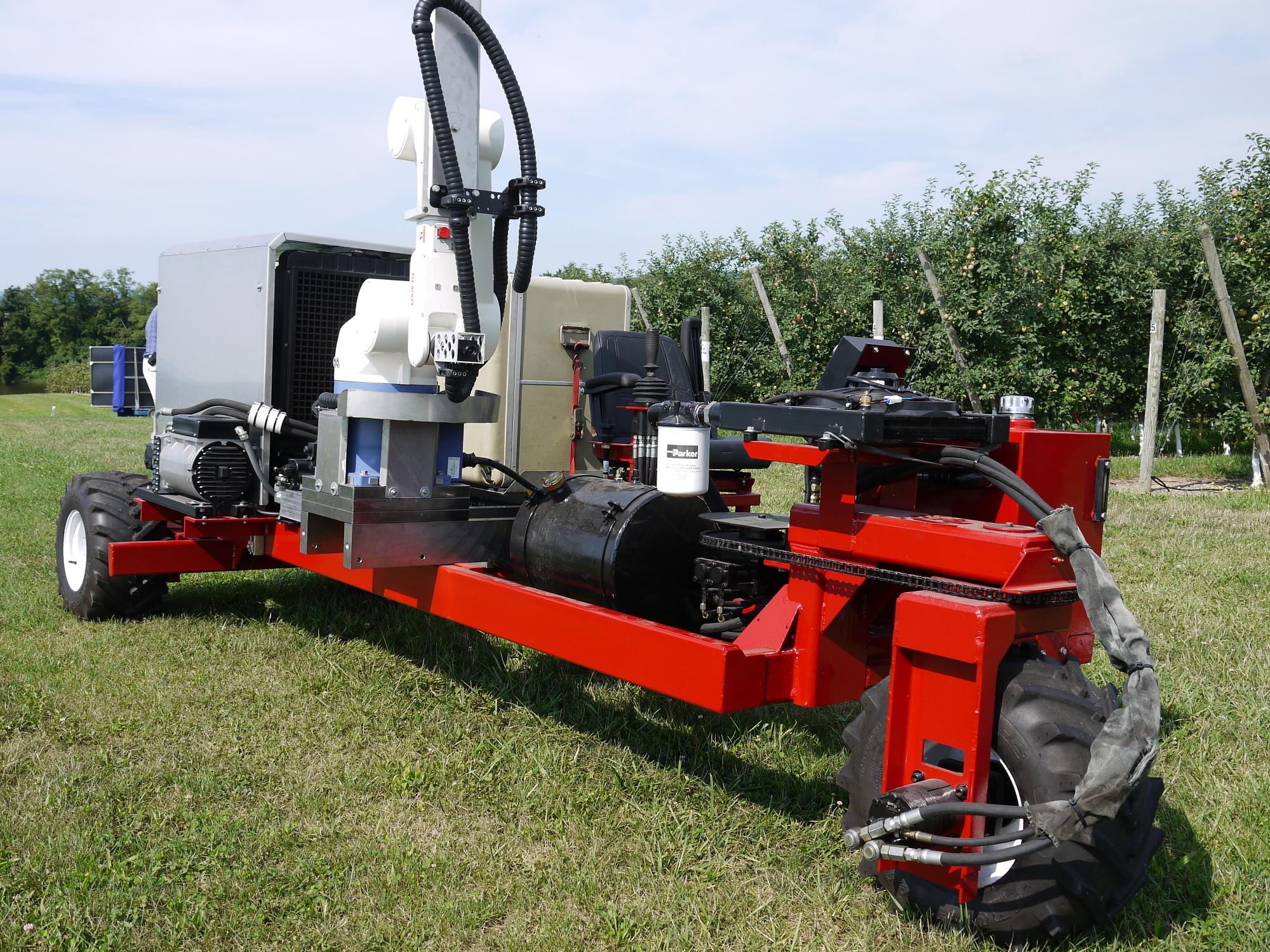}
\label{sf:large_robot}}
\caption{Two 6-axis robot arms used to generate the performance evaluation datasets.  In Subfigure \ref{sf:small_robot}, three cameras are mounted on the end-effector as in dataset 7, and this robot arm is rigidly attached to the laboratory floor.  Subfigure \ref{sf:large_robot} shows the larger robot arm mounted on a mobile platform with one camera (Subfigure \ref{sf:large_robot} image is courtesy of Edwin Winzeler).} 
\label{fig:DENSO_robots}
\end{figure}

The behavior of the robot-world hand-eye calibration methods is demonstrated on eight datasets in real laboratory and field settings.\footnote{{All of the datasets are available from \cite{tabb2017solving_dataset}.}}  These datasets represent different combinations of robots, cameras, and lenses; some of the datasets have multiple cameras.   Descriptions of the datasets are given in Table \ref{table:dataset_descriptions}.  The table also lists the number of robot positions ($n$) used for the calibration and the $rrmse$ error (in pixels) from the camera calibration step using Zhang's method  \cite{zhang2000flexible} to estimate the matrices $\mathbf{A}_i$.  For datasets 7 and 8, the third camera is a commodity RGB-D (Red-Green-Blue-Depth) camera; we calibrated only the color camera in this work.  Figure \ref{fig:example_images} shows the arrangement of cameras in dataset 7 as well as sample images from one position of the robot.  Figure \ref{fig:DENSO_robots} shows both of the robots used in this work; two 6-axis robot arms were used, a Denso VS-6577GM-B robot arm rigidly mounted to the floor and a Denso VM-60BIG robot arm mounted on a mobile platform.

In the rest of this section, we discuss some of the implementation details of our collection of proposed methods, and then discuss the error metrics used in reporting the performance results of the comparison of the proposed methods. 
\begin{table}[H]
\caption{Dataset descriptions. The number of robot positions is $n$, and $rrmse$ is the reprojection root mean square error from the camera calibration step.}
\begin{center}
\resizebox{\linewidth}{!}
{
\begin{tabular}{|c|c|c|c|c|c|c|}
\hline
Dataset & Image Size & Lens focal length & Robot & $n$ & $rrmse$ [pixels]\\
\hline
1 &   (640 $\times$ 480) & 8mm & Denso VS-6577GM-B & 88 & 0.185242\\
\hline
2 &   (2456 $\times$ 2058) & 8mm & Denso VS-6577GM-B & 28 & 0.199418\\
\hline
3 &   (2456 $\times$ 2058) & 6mm & Denso VM-60BIG & 36 & 0.540056\\
\hline
4 &   (1600 $\times$ 1200) & 6mm & Denso VM-60BIG & 20 & 0.447463\\
\hline
5 &   (1228 $\times$ 1029) & 6mm &Denso VS-6577GM-B & 15 & 0.124774\\
\hline
6 &    \parbox[t]{2cm}{(1228 $\times$ 1029)\\(1228 $\times$ 1029)} & \parbox[t]{2cm}{6mm\\6mm} & Denso VS-6577GM-B & 15 & \parbox[t]{1cm}{0.124774\\0.118215}\\
\hline
7 &   \parbox[t]{2cm}{(1228 $\times$ 1029)\\(1228 $\times$ 1029)\\(640 $\times$ 480)} & \parbox[t]{2cm}{6mm\\6mm\\not provided} & Denso VS-6577GM-B & 42 & \parbox[t]{1cm}{0.131076\\ 0.121817\\0.109997}\\
\hline
8 &   \parbox[t]{2cm}{(1228 $\times$ 1029)\\(1228 $\times$ 1029)\\(640 $\times$ 480)} & \parbox[t]{2cm}{8mm\\8mm\\not provided} & Denso VS-6577GM-B & 42 & \parbox[t]{1cm}{0.143151\\0.139296\\0.113589}\\
\hline
\end{tabular}
}
\end{center}
\label{table:dataset_descriptions}
\end{table}  

\vspace{-0.75cm}
\subsection{Implementation details}
\label{ss:implementation}

For the two proposed classes of cost functions, we use the implementation of the {Levenberg-Marquardt} method found in the software package \textit{Ceres} \cite{ceressolver}.  All of the results shown in this paper were generated on a workstation with one 12-core processor and 192 GB of RAM. The camera calibration was carried out using the OpenCV library's camera calibration functions \cite{opencv_library}.

For the first class of cost functions using simultaneous and separable versions of $c_1$ and $c_2$, the initial solutions of $\mathbf{p}_X$ and $\mathbf{p}_Z$ are set such that the corresponding directed cosine matrix is a $3 \times 3$ identity matrix, and the translation components are set to zero.\footnote{{Various different initial solutions were tested, and there was small or negligible difference in the solution quality versus using an identity matrix for rotation matrices and translation component with all elements zero. For this reason, we conclude that for the experiments covered in this paper, the first class of methods is not sensitive to initial solutions.}}  As mentioned previously, the second class of cost functions $rp_1$ and $rp_2$ are sensitive to initial solutions. To find an approximate solution to the minimization problem of $rp_1$ method that was shown in Equation \ref{eq:rsse_rwhec}, we used the simultaneous solution using the $c_2$ cost function (problem specified in Equation \ref{eq:c2_param_simultaneous}) as an initial solution.  To find an approximate solution to the problem of the $rp_2$ method shown in Equation \ref{eq:rsse_rwhec_with_k}, the solution from Equation \ref{eq:rsse_rwhec} was used as the initial solution. 

To present a comparative study against our collection of cost functions and methods, we implemented seven comparison methods: Zhuang \textit{et al.} \cite{Zhuang1994simultaneous}, Dornaika and Horaud \cite{dornaika1998simultaneous}: closed form and iterative, Hirsh \textit{et al.} \cite{hirsh2001iterative} and Li \textit{et al.} \cite{li2010simultaneous} based on the dual quaternion and Kronecker product, and the work of Shah \cite{Shah2013Solving}. Concerning the implementation details of Li \textit{et al.} \cite{li2010simultaneous} methods, we draw the reader's attention that the computation of coefficients $\lambda_1$ and $\lambda_2$ in the dual quaternion method was not specified clearly by the authors, thus we used a {Levenberg-Marquardt} method to find those values.  For the conversion of rotation matrices found by the Kronecker product method \cite{li2010simultaneous} into orthonormal matrices, we used the singular value decomposition method.

\subsection{Error metrics}
\label{subsec:Experiments-ErrorMetrics}

In presenting the comparison results of both of our collection of methods and those methods of others, we used the following error metrics: two types of mean rotation error, the mean translation error, the mean combined rotation and translation error, and the reprojection root mean squared error. Since we are interested in using our methods for reconstruction and vision tasks as indicated in Subsection \ref{section:contributions}, we also introduced a sixth metric related to reconstruction accuracy.  The following subsections describe each error type. 

\subsubsection{The Mean Rotation Error}
We list two rotation errors.  The first, which we denote Rotation Error 1 ($e_{R1}$), is derived from Equation \ref{eq:rotation_parts} and shown in Equation \ref{eq:rotError1}.
\begin{equation}
e_{R1} = \frac{1}{n}\sum_{i=0}^{n-1} || \mathbf{R}_{A_i} \mathbf{R}_X - \mathbf{R}_Z \mathbf{R}_{B_i} ||^2_{{F}}
\label{eq:rotError1}
\end{equation}

However, the $e_{R1}$ value may not be particularly meaningful when evaluating the methods, and has no units.  For this reason, we have a second rotation error, $e_{R2}$, that is representative of the difference between the left hand and right hand sides of Equation \ref{eq:rotation_parts}.  We find the relative rotation between the two sides (Equation \ref{eq:relative_transformation}) and then compute the angle of the relative rotation using the axis-angle representation, which is represented as $angle()$ in Equation \ref{eq:rotError2}; in the results shown in Section \ref{section:Res}, its units are degrees.

\begin{align}
\mathbf{R}_{i} =  (\mathbf{R}_Z \mathbf{R}_{B_i})^T (\mathbf{R}_{A_i} \mathbf{R}_X)\label{eq:relative_transformation}\\
e_{R2} = \frac{1}{n}\sum_{i=0}^{n-1} angle(\mathbf{R}_{i})\label{eq:rotError2}
\end{align}

\subsubsection{The Mean Translation Error} \label{ss:mean_trans}
The translation error is derived from Equation \ref{eq:translation_parts} and shown in Equation \ref{eq:transError}, where the units are in millimeters to be consistent with our selection of units for camera calibration error. 
\begin{equation}
e_t = \frac{1}{n}\sum_i || (\mathbf{R}_{A_i} \mathbf{t}_X + \mathbf{t}_{A_i}) - (\mathbf{R}_{Z_i} \mathbf{t}_{B} + \mathbf{t}_{Z_i}) ||^2
\label{eq:transError}
\end{equation}

\subsubsection{The Mean Combined Rotation and Translation Error}
This error is shown in Equation \ref{eq:RotTransError} and has no units.
\begin{equation}
e_C = \frac{1}{n}\sum_i || \mathbf{A}_i \mathbf{X} - \mathbf{Z} \mathbf{B}_i||^2_{{F}}
\label{eq:RotTransError}
\end{equation}

\subsubsection{The Reprojection Root Mean Squared Error (rrmse)} 
This error is shown in Equation \ref{eq:mse_robot} and its units are pixels.
\begin{equation}
rrmse= \sqrt{\frac{1}{mn}\sum_{i = 0}^{n-1}\sum_{j=0}^{m-1} ||\vec{\mathbf{x}}_{ij}- f({\mathbf{k}}, \left[\mathbf{Z}\mathbf{B}_i{\mathbf{X}^{-1}}\right]_{3\times 4} \vec{\mathcal{X}}_j)||^{2}}
\label{eq:mse_robot}
\end{equation}

\subsubsection{The Reconstruction Accuracy Error}

Since we are also interested in reconstruction accuracy when using a robot to acquire images with associated camera calibration information, our final metric aims to represent the reconstruction accuracy.

The idea behind this metric is given correspondences from different images acquired at different robot positions, estimate the world points that generated those correspondences, and compute the difference between the estimated versus true world points to represent reconstruction accuracy.  We borrow some of the notation from Section \ref{ss:ReprojectionError}: $\vec{\mathbf{x}}_{ij}$ is the $j$th image point from the $i$th robot position and corresponds to the three-dimensional point $\vec{\mathcal{X}}_j$.  First, we estimate the most likely three-dimensional point (represented by $\hat{\mathcal{Y}}_j$) that generated the $n$ $\vec{\mathbf{x}}_{ij}$ image points by solving the minimization problem in Equation \ref{eq:min_world_point}. 


\begin{equation}
\hat{\mathcal{Y}}_j = \argmin\limits_{\mathcal{Y}_j} \sum_{i = 0}^{n-1} ||\vec{\mathbf{x}}_{ij} -  f(\mathbf{k}, \left[\mathbf{Z}\mathbf{B}_i{\mathbf{X}^{-1}}\right]_{3\times 4}\mathcal{Y}_j)||^2 
\label{eq:min_world_point}
\end{equation}

Then, the reconstruction accuracy error ($rae$) is the average Euclidean distance between the estimated $\hat{\mathcal{Y}}_j$ points and calibration object points $\vec{\mathcal{X}}_j$:
\begin{equation}
rae = \frac{1}{m} \sum_{j = 0}^{m-1} || \hat{\mathcal{Y}}_j - \vec{\mathcal{X}}_j ||^2
\label{eq:rae}
\end{equation}


\section{{Performance Evaluation on simulated datasets}} \label{section:SimulatedExperiments-ErrorMetrics}

The comparison methods as well as a subset of our proposed methods are evaluated on simulated datasets. For the simulated datasets contained in this paper, only $\mathbf{A}_i$, $\mathbf{B}_i$, $\mathbf{X}$, and $\mathbf{Z}$ are generated.  As a result, we do not evaluate our proposed second class of methods, since doing so would also require simulating camera models. The protocol for generating the simulated datasets closely follows that of Shah 2013 \cite{Shah2013Solving}, Section 5.1. Briefly, the rotation components of $\mathbf{A}_i$, $\mathbf{X}$, and $\mathbf{Z}$ are set as random rotation matrices, and the translation components are drawn from the standard uniform distribution $(0, 1)$ using a random number generator, for $i = 1, 2, ..., 25$. Then, the homogeneous matrix $\mathbf{B}_i$ is computed as: 

\begin{equation}
\mathbf{B}_i = \mathbf{Z}^{-1} \mathbf{A}_i \mathbf{X}
\label{eq:b_i}
\end{equation}

Noise is then introduced to $\mathbf{B}_i$, only in the rotation component, by converting the rotation matrix to the quaternion representation and adding random noise, and then converting back to the matrix representation. $\eta$ is the parameter controlling the magnitude of the random noise added to $\mathbf{B}_i$. $\eta \in (0, 0.25]$ and is evenly spaced over 19 values along that interval.  Data was generated for ten trials of the experiment.

The protocol above sets the translation components within the interval of $(0, 1)$, which reflects a bias towards a particular use of units.  Depending on the particular experiment setup and choices made, this interval may or may not be representative of that experimental setup.  Consequently, we denote the simulated datasets using Shah's protocol \cite{Shah2013Solving} as Simulated Dataset I. We generated another dataset using the same protocol, except that the translation components were within the interval of $(0, 1000)$, and refer to this as Simulated Dataset II.  We chose the interval $(0, 1000)$ since in our real experiments we use millimeters for translation components to be consistent with camera calibration; of course, the use of meters would be another choice and this choice is reflected in Simulated Dataset I.

\subsection{Error metrics for simulated datasets}
\label{subsec:Experiments-ErrorMetrics-simulated}

Since the ground truth $\mathbf{X}$ and $\mathbf{Z}$ are known for the simulated datasets, for each calibration method the difference between the rotation and translation components are computed.  Again, we follow closely Shah's presentation \cite{Shah2013Solving} in our description of the error metrics for the simulated experiments.

\subsubsection{Rotation error} \label{ss:rot_error_simulation}
Let $\hat{\mathbf{R}}_X$ be the rotation matrix estimated by a calibration method for a simulated dataset, and ${\mathbf{R}}_X$ be the ground truth rotation matrix for  $\mathbf{X}$.  Then, the error in the rotation component is ${e}_{RX}$:

\begin{equation}
{e}_{RX} = || \hat{\mathbf{R}}_X - \mathbf{R}_X ||_F
\label{eq:r_X}
\end{equation}
and the same formula follows to compute the rotation error for $\mathbf{Z}$, $e_{RZ}$.

\subsubsection{Translation error}\label{ss:trans_error_simulation}

Let $\hat{\mathbf{t}}_X$ be the translation vector estimated by a calibration method for a simulated dataset, and ${\mathbf{t}}_X$ be the ground truth translation vector for $\mathbf{X}$.  Then, the error in the translation component is $e_{tX}$:

\begin{equation}
e_{tX} = || \hat{\mathbf{t}}_X - \mathbf{t}_X ||
\label{eq:t_X}
\end{equation}
and the same formula follows to compute the translation error for $\mathbf{Z}$, $e_{tZ}$.

\section{Experimental Results}
\label{section:Res}
In this section, we first show and discuss comparison results between our methods and the seven methods we referred to in the previous section on real datasets.  Then, we show and discuss comparison results using the two simulated datasets.

Tables \ref{table:dataset1} through \ref{table:dataset8} show the results using the five error metrics described in Subsection \ref{subsec:Experiments-ErrorMetrics} and the eight real datasets described in Table \ref{table:dataset_descriptions}.\footnote{While use of tables is perhaps not the easiest for the reader, we note that the huge range of values for each of the {six} error metrics made other types of representation (graphs, etc.) infeasible.}  These comparison results also include the running time for our implementation of each method in seconds.

In Tables \ref{table:dataset1} -- \ref{table:dataset5}, the first seven rows correspond to the seven comparison methods from the literature, listed in chronological order: Zhuang \textit{et al.} \cite{Zhuang1994simultaneous}, Dornaika and Horaud \cite{dornaika1998simultaneous}: closed form and iterative, Hirsh \textit{et al.} \cite{hirsh2001iterative} and Li \textit{et al.} \cite{li2010simultaneous} based on the dual quaternion and Kronecker product, and the method of Shah \cite{Shah2013Solving}, whereas the remainder of the rows correspond to our proposed methods.\footnote{Only the first author is mentioned in the tables' text for brevity.} The comparison methods were not evaluated on Datasets 6, 7, and 8 and are not shown in Tables \ref{table:dataset6} - \ref{table:dataset8}, because those datasets have more than one camera, and to the best of our knowledge, there are currently no other robot-world hand-eye calibration methods for multiple cameras.

\begin{figure*}
\centering
	\includegraphics[ width=0.70\linewidth]{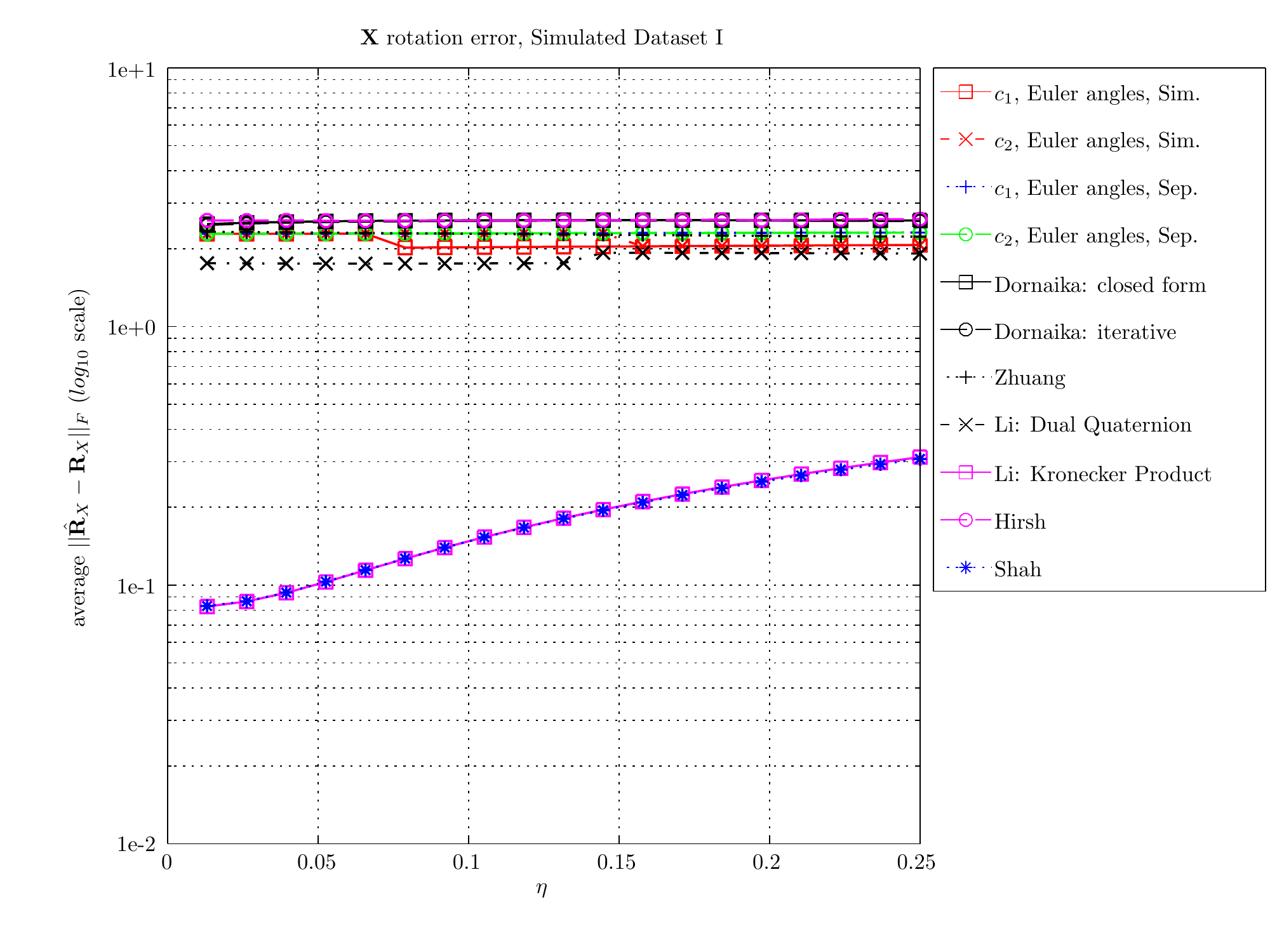}

	\includegraphics[ width=0.70\linewidth]{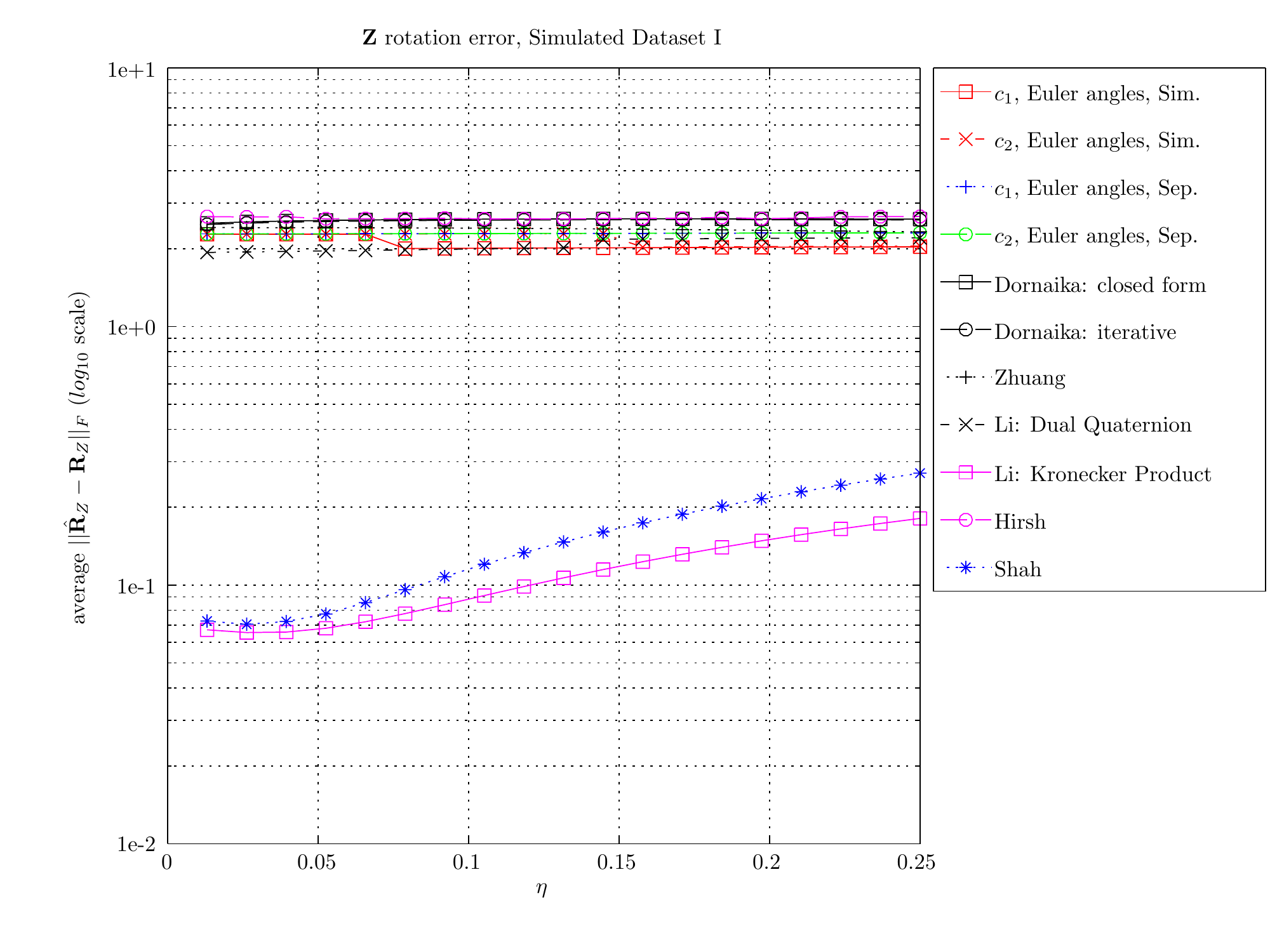}
\caption{\textbf{Best viewed in color.} {Rotation error results for Simulated Dataset I, as compared to the ground truth $\mathbf{X}$ and $\mathbf{Z}$, using the four metrics of Subsection \ref{subsec:Experiments-ErrorMetrics-simulated}.} }
\label{fig:simulated_DSIA}
\end{figure*}

\begin{figure*}
\centering
	\includegraphics[ width=0.70\linewidth]{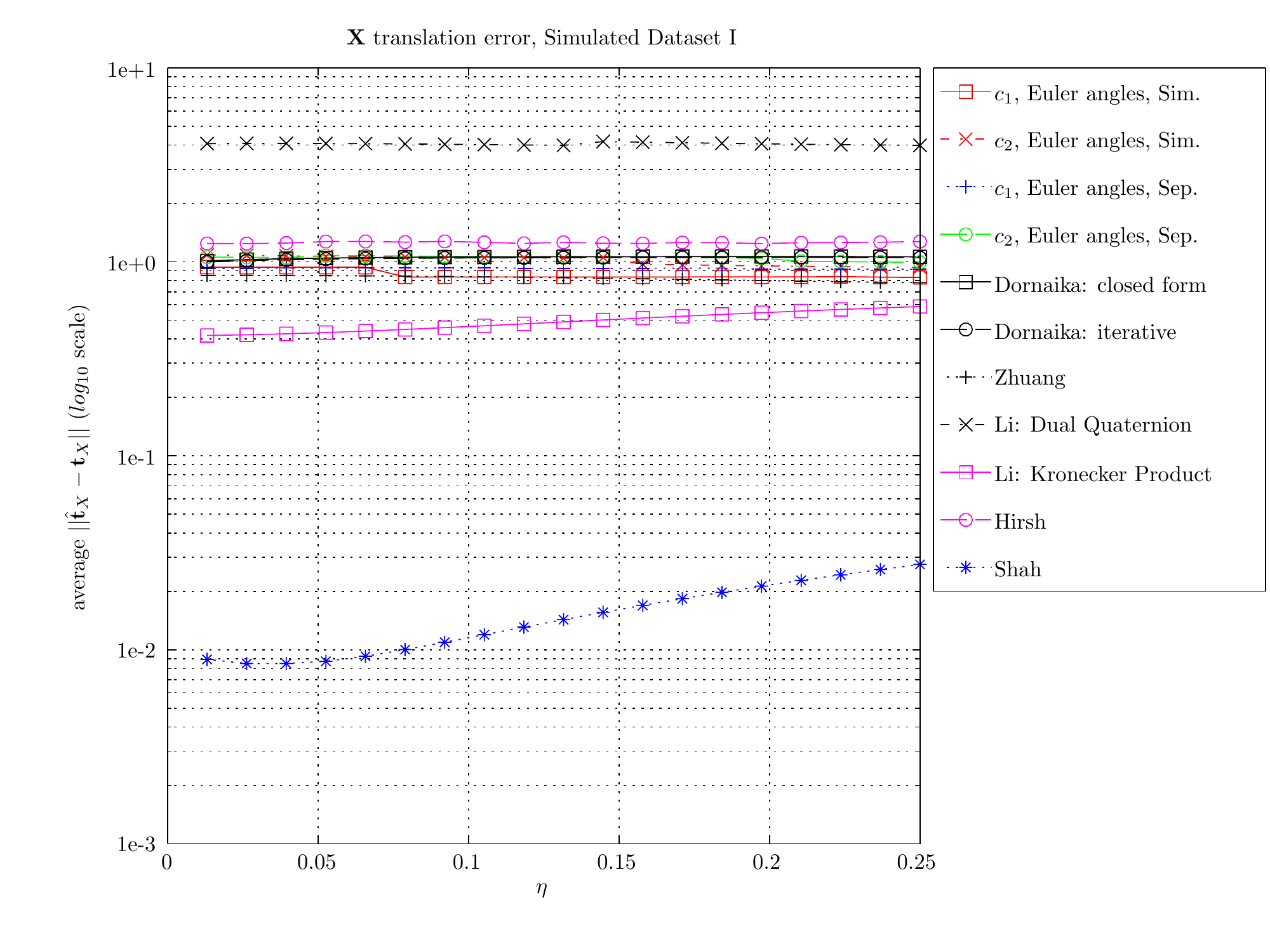}

	\includegraphics[ width=0.70\linewidth]{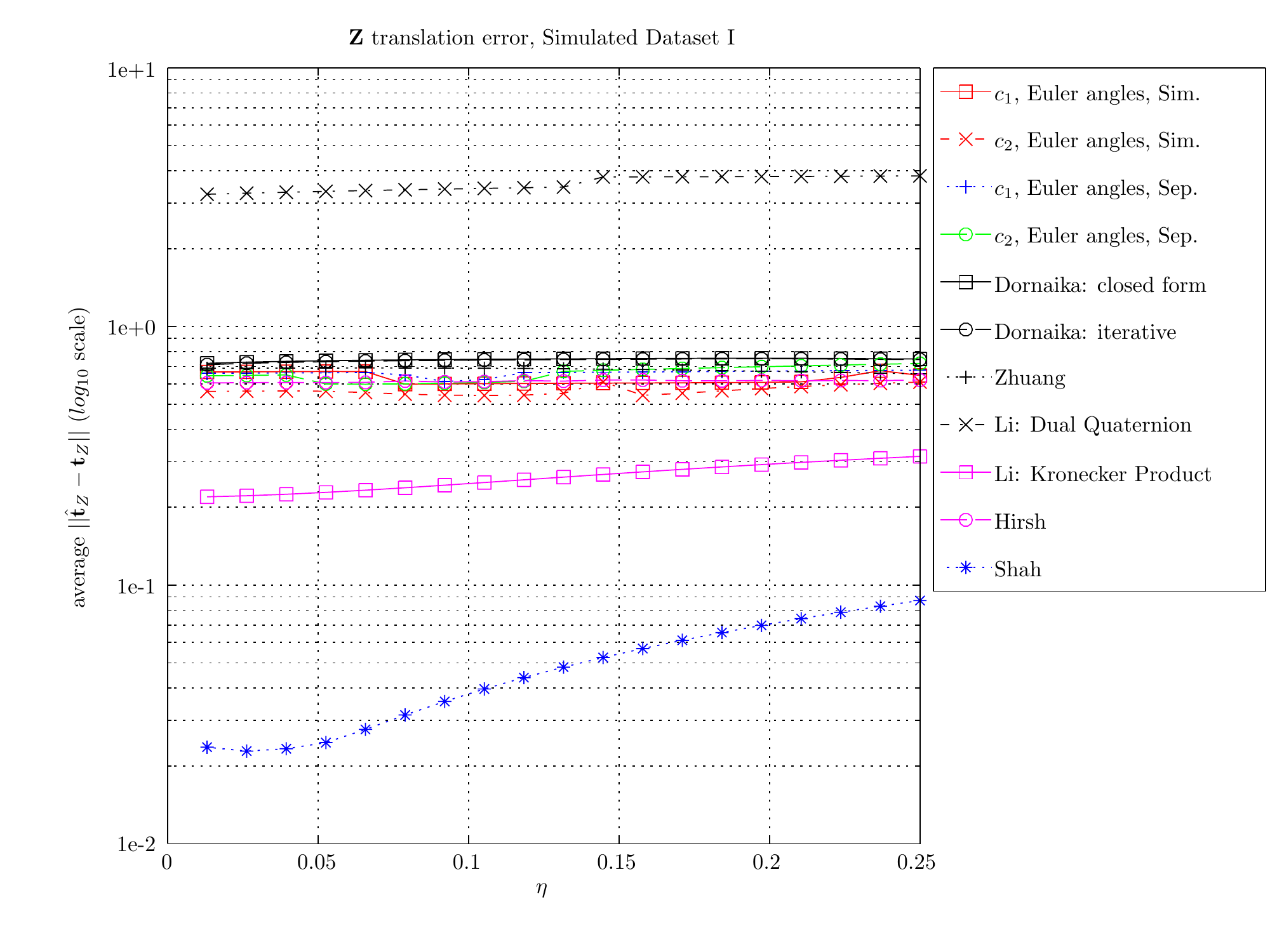}
\caption{\textbf{Best viewed in color.} {Translation error results for Simulated Dataset I, as compared to the ground truth $\mathbf{X}$ and $\mathbf{Z}$, using the four metrics of Subsection \ref{subsec:Experiments-ErrorMetrics-simulated}.} }
\label{fig:simulated_DSIB}
\end{figure*}

\begin{figure*}
\centering
	\includegraphics[ width=0.70\linewidth]{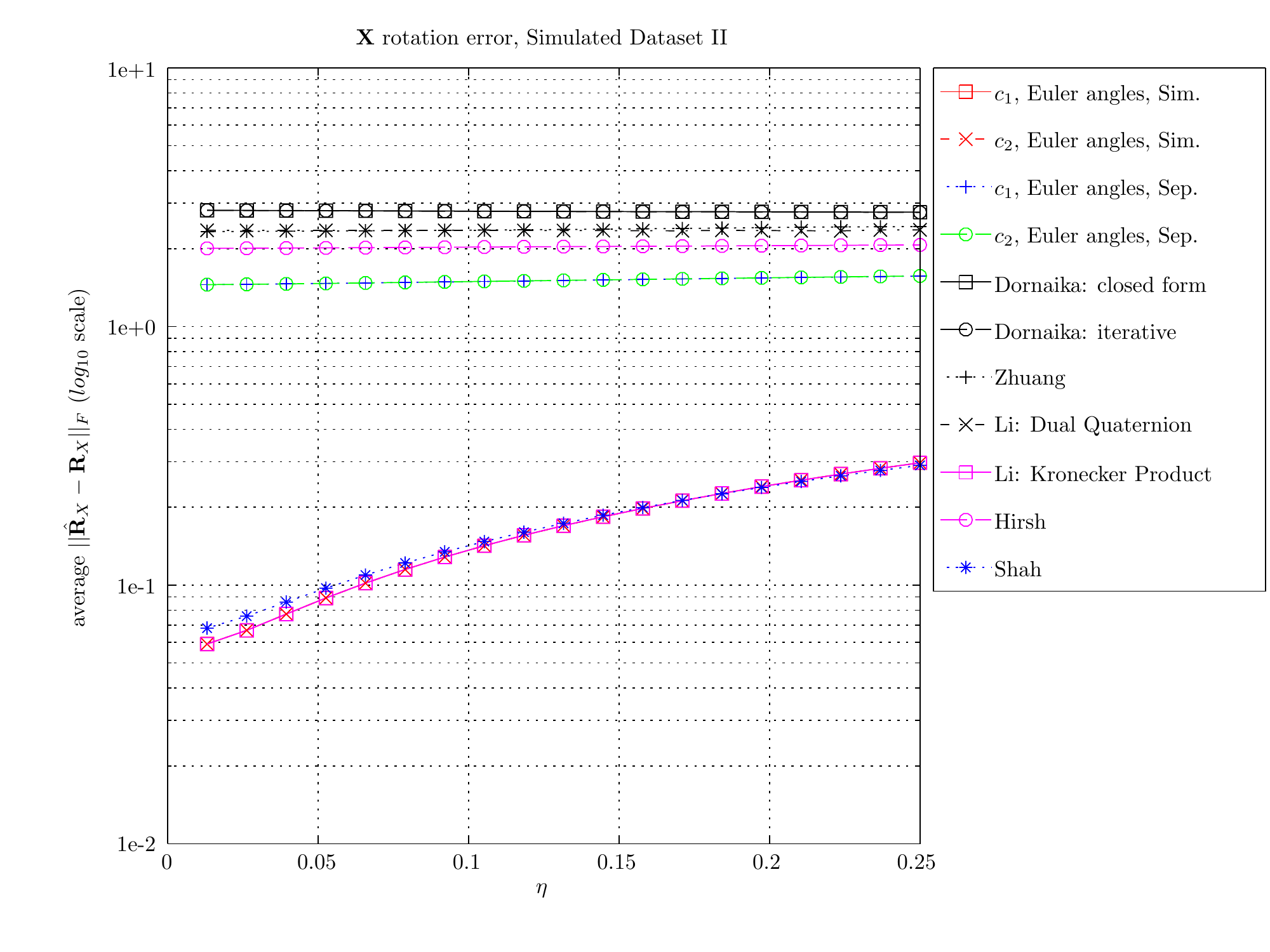}

	\includegraphics[ width=0.70\linewidth]{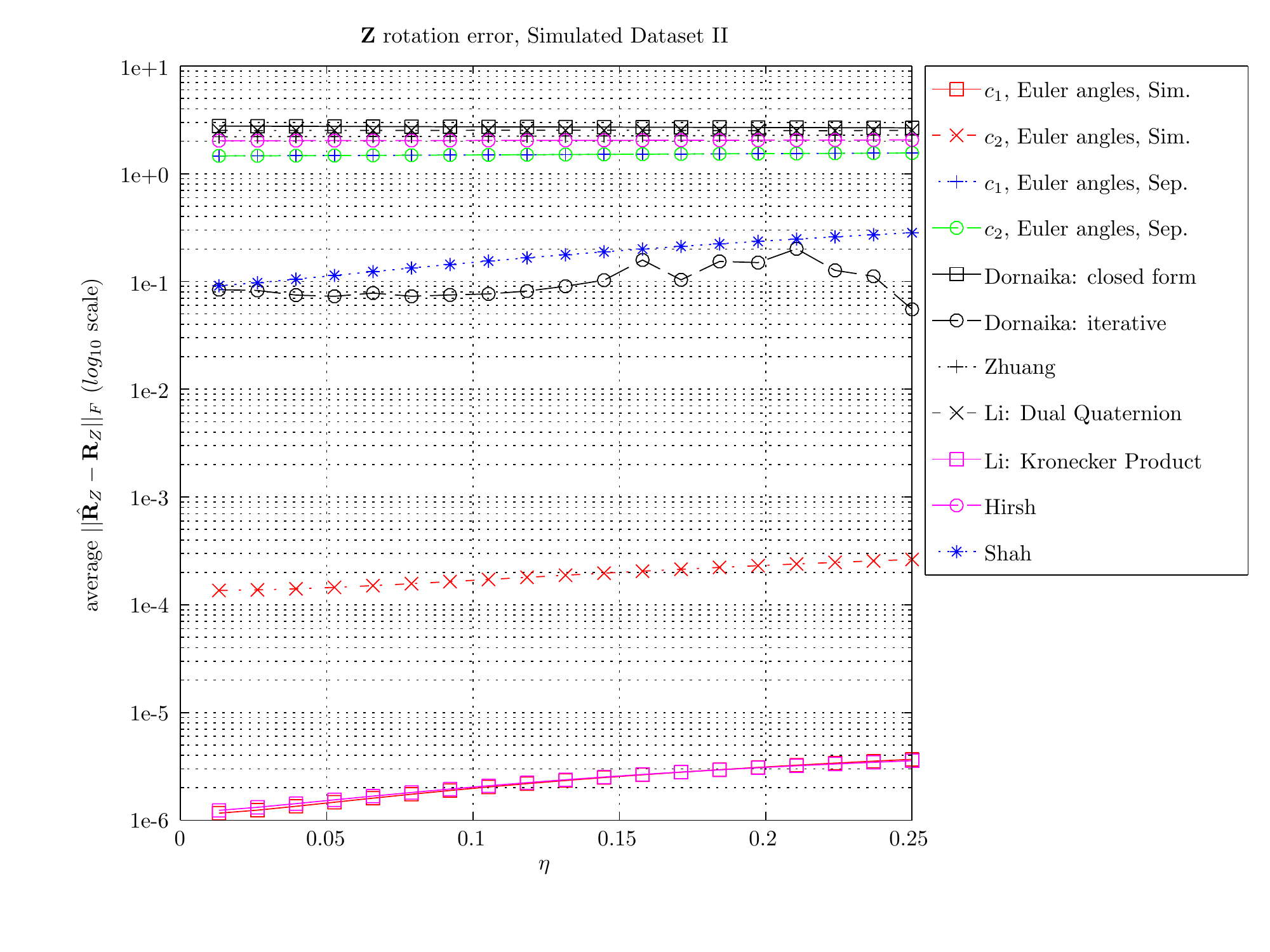}
\caption{\textbf{Best viewed in color.} {Rotation error results for Simulated Dataset II, as compared to the ground truth $\mathbf{X}$ and $\mathbf{Z}$, using the four metrics of Subsection \ref{subsec:Experiments-ErrorMetrics-simulated}.} }
\label{fig:simulated_DSIIA}
\end{figure*}

\begin{figure*}
\centering
	\includegraphics[ width=0.70\linewidth]{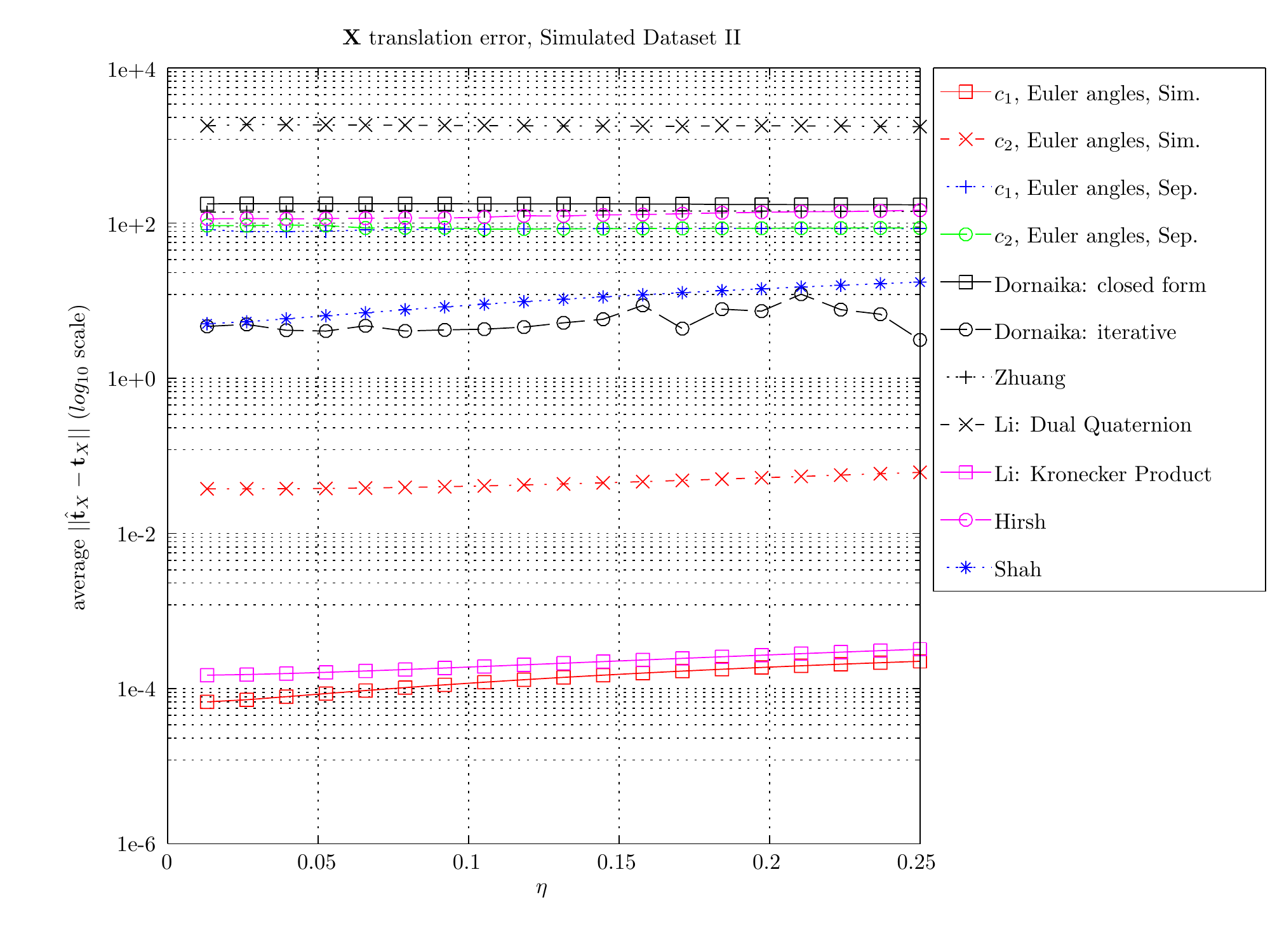}

	\includegraphics[ width=0.70\linewidth]{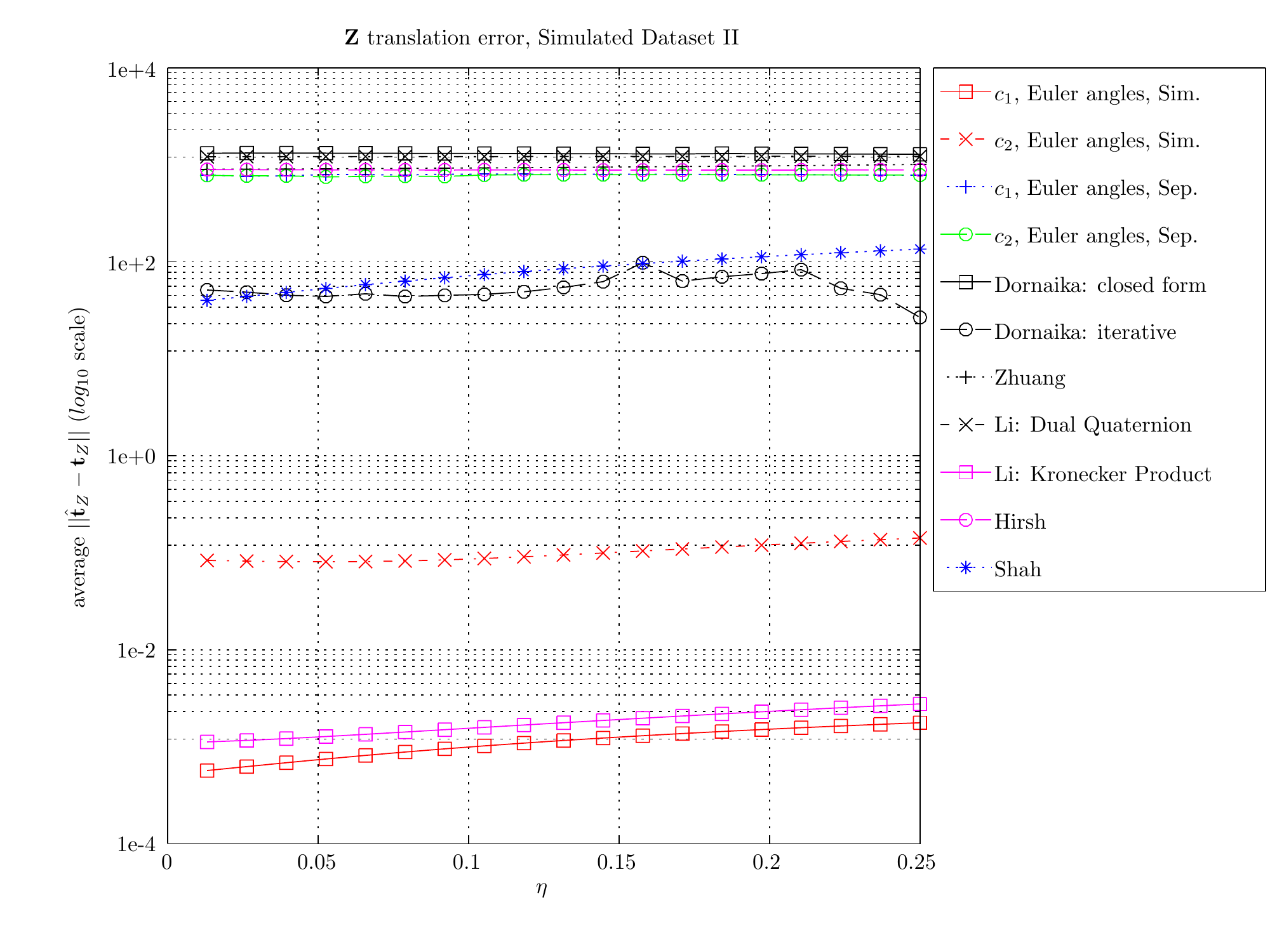}

\caption{\textbf{Best viewed in color.} {Translation error results for Simulated Dataset II, as compared to the ground truth $\mathbf{X}$ and $\mathbf{Z}$, using the four metrics of Subsection \ref{subsec:Experiments-ErrorMetrics-simulated}.} }
\label{fig:simulated_DSIIB}
\end{figure*}

Results for Simulated Datasets I and II are shown graphically in Figures \ref{fig:simulated_DSIA} - \ref{fig:simulated_DSIIB}. The seven comparison methods were evaluated as well as the first class from our proposed collection of methods. Within the first class of our proposed collection of methods, the results were very similar for different choices of rotation parameterization.  Consequently, we only listed one rotation parameterization choice, that of Euler angles, to allow better readability. 

\subsection{Discussion of comparison methods on real datasets}
We mention here that the Dornaika and Horaud iterative method \cite{dornaika1998simultaneous} does not converge for any datasets using the \textit{Ceres} solver \cite{ceressolver}, where derivatives are automatically computed by the software.  In comparison, in our previous work \cite{Tabb2015Parameterizations}, where we used the \textit{levmar} solver \cite{lourakis04LM}, the Dornaika and Horaud iterative method converged after only 2-3 iterations after termination conditions were met, however, the results were very similar to that produced with the authors' closed form method (which served as the iterative method's initial solution). In summary, the behavior of different solvers may give different results, but the main conclusion we have made about this method is that the large penalty terms used to enforce orthonormality of the rotation matrices result in either very little change from the initial solution, or non-convergent behavior.  

Another similarity among the comparison methods is the Hirsh \textit{et al.} \cite{hirsh2001iterative} and Shah \cite{Shah2013Solving} methods. Both of these methods consistently produce the lowest values of rotation errors $e_{R1}$ and $e_{R2}$ that are usually the same up to 5 digits of precision. In contrast, the Dornaika and Horaud closed-form method \cite{dornaika1998simultaneous} usually has one of the highest rotation errors. With regard to the mean translation error ($e_t$) and the combined rotation and translation error ($e_C$), Shah's method in general has a good performance relative to other comparisons methods as it results in one of the lowest values among the datasets until Dataset 5 (Table \ref{table:dataset5}). With respect to the reprojection mean square error ($rrmse$), the Hirsh \textit{et al.} \cite{hirsh2001iterative} and Shah \cite{Shah2013Solving} methods both perform generally the best, with Shah's method resulting in lower $rrmse$ in general as compared to the method of Hirsh \textit{et al.}, though this relationship is inverted in Datasets 1 and 5. For example, the method of Hirsh \textit{et al.} has the lowest $rrmse$ in Datasets 1 ($3.68829$ pixels), but large $rrmse$ in Dataset 5 ($238.041$ pixels), though still lower compared to Shah's method's $rrmse$ value ($439.512$ pixels).

The first conclusion drawn from Tables \ref{table:dataset1} -- \ref{table:dataset5}, which correspond to datasets with a single camera, is that the behavior of many methods is dataset dependent.  This dataset dependency may result from many factors, including the positions chosen for calibration, the robot model, and the way in which the robot is mounted, which may introduce new errors.  Recall from Section \ref{section:Experiments-ErrorMetrics} that the DENSO VM-60BIG from Datasets 3 and 4 is mounted on a mobile unit with tires, so as the robot moves, there might be some movement in the robot base.

The dataset dependency of the comparison methods is illustrated with a selection of examples.  For instance, the combined rotation and translation error ($e_C$) of Zhuang \textit{et al.} \cite{Zhuang1994simultaneous} is the lowest of the group of comparison methods in Dataset 1 ($233.3$), in the middle range in Dataset 2, and greater than $170,000$ in Datasets 3, 4, and 5.  The Kronecker product method from Li \textit{et al.} \cite{li2010simultaneous} performed well relative to the comparison methods for Dataset 5 (Table \ref{table:dataset5}), where the combined rotation and translation error was $e_C = 329.24$ and the other methods had combined rotation and translation error $e_C \geq 13,000$.  However, for no other single dataset that has one camera does the Li \textit{et al.} Kronecker product method have the lowest combined rotation and translation error relative to the other comparison methods.

\subsection{Discussion of the proposed collection of methods on real datasets}

\subsubsection{First class of cost functions}
Within the first class of cost functions, the methods based on minimizing $c_1$ ($i.e.$ simultaneous and separable under different choices of parameterization of the rotational components) tend to have a lower value of combined rotation and translation error ($e_C$), while methods based on minimizing $c_2$ have lower values of $rrmse$ as compared to $c_1$ methods.  Exceptions are Datasets 7 and 8, which have very similar values of $e_C$ and $rrmse$ for $c_1$ and $c_2$ separable versions (see Tables \ref{table:dataset7} and \ref{table:dataset8}).   

Specifically, within the first class of cost functions, the separable versions of the methods tended to have lower rotation errors $e_{R1}$, $e_{R2}$, but higher translation errors $e_t$ and combined rotation and translation error $e_C$, as well as reprojection error $rrmse$.  This is particularly obvious in Dataset 5 (Table \ref{table:dataset5}), where the reprojection root mean square errors of the separable versions are all greater than $48$ pixels, while the simultaneous versions have a maximum $rrmse$ of $3.89$ pixels.   The separable methods have a much greater $rrmse$ than the simultaneous versions in particular in Dataset 6, Table \ref{table:dataset6}.

\subsubsection{Second class of cost functions}

As expected when using reprojection error as a cost function versus the relationship $\mathbf{AX}= \mathbf{ZB}$ (and its variations), the second class of methods consistently gives results with lower values of $rrmse$ than the first class of methods or the comparison methods. On the other hand, the rotation errors, translation errors and combined rotation and translation errors, when using the second class of methods, are greater than the first class of methods as well as some of the comparison methods. In general, the change in $rrmse$ using $rp_1$ (where intrinsic camera calibration parameters are treated as constants) versus $rp_2$ (where intrinsic camera calibration values are parameters) is small.

\subsubsection{Effect of rotation representation}

It seems that the choice of rotation representation -- Euler angles, axis-angle, or quaternion -- has only a small effect on the results, and those differences may be dependent on the particular solver used.  For instance, in our prior work \cite{Tabb2015Parameterizations}, the simultaneous version of the cost function $c_1$ was used based on the Euler angles parameterization (in that work, labeled as Euler Parameterization I) using the \textit{levmar} solver \cite{lourakis04LM} and a provided Jacobian matrix. In that work, when tested on Dataset 1, the $rrmse$ was  3.62709 pixels.  Whereas using the \textit{Ceres} solver \cite{ceressolver} with automatic Jacobian matrices for the same dataset, the $rrmse$ values were 12.13, 3.62686 and 3.62641 pixels, respectively for the Euler angles, axis-angle and quaternion versions (Table \ref{table:dataset1}).  This difference in $rrmse$ occurs even in the cases when the combined rotation and translation error ($e_C$) between the three rotation representations, using \textit{Ceres},  differs by only $0.001$.  Usually, the differences between the rotation representations on all of the metrics are relatively small, though exceptions are Datasets 5 and 6 (Tables \ref{table:dataset5} and \ref{table:dataset6}), the separable $c_1$ and $c_2$ methods.

\subsection{Reconstruction accuracy performance for the proposed collection of methods and comparison methods}
\label{subsec:rae}
While reprojection root mean square error is a good indicator of relative reconstruction accuracy error ($rae$), there is not a monotonic relationship of $rrmse$ to $rae$.  We observed that the $rp_2$ version from the second class in our collections returns slightly lower values of $rrmse$ than $rp_1$.  Concerning reconstruction accuracy error of the $rp_2$ method versus the $rp_1$ method, $rae$ increases slightly in datasets with better selections of robot positions (Datasets 1, 2, 3, 4, 7, 8) using $rp_1$ and decreases slightly in datasets with poorer selections of robot positions (Datasets 5 and 6) when using $rp_2$ method as compared to the $rp_1$ method. Concerning our first class of cost functions ($c_1$ and $c_2$), we observed that using the $c_2$ simultaneous method always gives a lower $rae$. In particular, the $c_2$ simultaneous version in Dataset 2 has $rae \leq 0.1$ mm for all three rotation representations, and this pattern ($rae \leq 0.1$ mm) is repeated in the separable versions of $c_2$ as compared to $c_1$ ($rae \leq 15$ mm, $rae \leq 2.38$ mm for separable and simultaneous, respectively). Additionally, it is observed that the separable versions of both $c_1$ and $c_2$ perform well in some datasets with respect to $rae$ and worse in others ($e.g.$ perform well in Dataset 2 ($rae \leq 2.381$ mm), and perform poorly in Dataset 5 ($rae \geq 33,338.9$ mm)).

Within the group of comparison methods, the methods of Hirsh \textit{et al.} \cite{hirsh2001iterative} and Shah \cite{Shah2013Solving} consistently have the lowest values of rotation error, and both of these methods perform generally well with respect to $rrmse$.  In general, Shah's computation of translation vectors results in lower translation error ($e_t$) and $rrmse$ as compared to the method of Hirsh \textit{et al.}, though this relationship is inverted in Dataset 5.  However, none of the comparison methods performs as well as our proposed $rp_1$ and $rp_2$ methods with respect to reconstruction accuracy error, though for some datasets the methods of Hirsh \textit{et al.} and Shah are comparable to the $rae$ of our first class of proposed methods.

\subsection{Discussion of the simulated experiments}

The examination of the Simulated Dataset I and II results offer some additional insights into the behavior of the comparison and proposed methods, since the ground truth $\mathbf{X}$ and $\mathbf{Z}$ are known.  The Simulated Dataset I results for the rotation error for $\mathbf{X}$ and $\mathbf{Z}$, show that the Shah and Li \textit{et al.} Kronecker product methods consistently producing the rotation matrices closest to the ground truth, while the rest of the methods have very similar error values ranging from $2$ and $2.6$.  The Shah and Li \textit{et al.} Kronecker product methods also have the lowest translation error for $\mathbf{X}$ and $\mathbf{Z}$ in Simulated Dataset I, while the rest of the methods have similar error values.  One exception is the Li \textit{et al.} dual quaternion method, which has the largest error than the rest of the methods for the translation components of $\mathbf{X}$ and $\mathbf{Z}$.

Simulated Dataset II is different from Simulated Dataset I in the scaling of the translation components, and this difference affects the performance of the methods.  Concerning rotation error, as before the Shah and Li \textit{et al.} Kronecker product methods produce the lowest values, but in Simulated Dataset II the simultaneous versions from our first class of proposed methods also produce comparable values.  For translation error, the Li \textit{et al.} Kronecker product and simultaneous versions from our first class of proposed methods produce very low error values ($\leq0.061$), while the remainder have values which vary considerably.

From these experiments, it is clear that some methods are sensitive to the scaling and distribution of the translation components, and a different method for dataset generation may yield different observations.  The Li \textit{et al.} Kronecker product method seems to be a robust choice in both of these settings, the Shah method for the Simulated Dataset I setting, and our first class of proposed methods for the Simulated Dataset II setting.

\subsection{Recommendations and Summary}

When choosing a robot-world, hand-eye calibration method to use, we hope that our experiments can aid researchers choose the method best suited to a particular task or application.  For instance, the methods of Hirsh \textit{et al.} \cite{hirsh2001iterative} and Shah \cite{Shah2013Solving}, and the $c_1$ and $c_2$ separable versions from the first class of our proposed methods consistently give the lowest rotation errors, whereas the $c_1$ simultaneous method returns the lowest combined rotation and translation error $e_C$.  For the best $rrmse$ and $rae$ without using camera reprojection error in the cost function, the $c_2$ simultaneous method is the best choice.  When using camera reprojection error as the cost function, $rp_1$ is generally better on $rae$ with a higher quality camera calibration and $rp_2$ is better than $rp_1$ with a poorer quality intrinsic camera calibration.  Within the collection of methods we propose, the choice of rotation representation does not greatly affect the results when using a modern solver such as Ceres \cite{ceressolver}. 

\begin{table*}
\caption{Comparison of methods using the error metrics described in Section \ref{subsec:Experiments-ErrorMetrics} for Dataset 1.}
\begin{center}
\resizebox{\linewidth}{!}
{
\begin{tabular}{|c|S[table-format=3.4]|S[table-format=1.10]|S[table-format=2.6]|S[table-format=2.6]|S[table-format=6.3]|S[table-format=3.6]|S[table-format=8.6]|}
\hline
\textbf{Method} & \textbf{Time (s)} & \textbf{$e_{R1}$} & \textbf{$e_{R2}$}(degrees) &\textbf{$e_t$}(mm) & \textbf{$e_C$} & \textbf{$rrmse$}(pixels) & \textbf{$rac$}(mm)\\
\hline
Zhuang \cite{Zhuang1994simultaneous} & 0.025 & 0.000101190  &   0.336407 & 1.62805 &  233.3  & 4.35341 & 434.012 \\
\hline
Dornaika \cite{dornaika1998simultaneous}: Closed Form & 0.01 &  0.261051    &  19.5162 &  61.24160 &   330047 &  762.333   & 195437.640\\
\hline
Dornaika \cite{dornaika1998simultaneous}: Iterative & 3.464 & 1.947 & 58.652 & 56.7113 & 283025   &  272.697 & 231374732 \\
\hline
Hirsh \cite{hirsh2001iterative} & 2.273 & 0.000098372  & 0.331641 &  2.02068  &  359.3 &  3.68820 &  189.545\\
\hline
Li \cite{li2010simultaneous}: Dual Quaternion & 0.494 & 0.000174349  &  0.506214 & 1.90747 &    320.2 &  4.30692 & 103.749\\
\hline
Li \cite{li2010simultaneous}: Kronecker product & 0.005 & 0.000102497 & 0.347818  & 2.25149  &  446.1  &  5.41810 & 388.55\\
\hline
Shah \cite{Shah2013Solving} & 0.001 & 0.000098372  & 0.331641 &  1.63616 &  235.577 & 4.10190 & 403.674\\
\hline
\end{tabular}
}
\end{center}
\label{table:dataset1}
\end{table*}
  
\begin{table*}
\caption{Comparison of methods using the error metrics described in Section \ref{subsec:Experiments-ErrorMetrics} for Dataset 2.}
\begin{center}
\resizebox{\linewidth}{!}
{
\begin{tabular}{|c|S[table-format=3.4]|S[table-format=1.10]|S[table-format=2.6]|S[table-format=2.6]|S[table-format=6.3]|S[table-format=3.6]|S[table-format=8.6]|}
\hline
\textbf{Method}  & \textbf{Time (s)} & \textbf{$e_{R1}$} & \textbf{$e_{R2}$} (degrees) &\textbf{$e_t$}(mm) & \textbf{$e_C$} & \textbf{$rrmse$}(pixels) & \textbf{$rae$}(mm)\\
\hline
Zhuang \cite{Zhuang1994simultaneous} &  0.014 & 0.0000625721 &            0.288076                & 1.307             & 47.8309                    & 11.8458 & 52.65020\\
\hline
Dornaika \cite{dornaika1998simultaneous}: Closed Form & 0.003 &  0.404248     &        24.1399     &        79.3305 &              176214 &    2959.58 & 5580067.998  \\
\hline
Dornaika \cite{dornaika1998simultaneous}: Iterative & 1.206 & 1.30484 &             45.2478 &             74.8997 &              157080         &         1329.87 & 173943689.69\\
\hline
Hirsh \cite{hirsh2001iterative} & 0.35 & 0.0000199547 &            0.149794 &            0.888279 &             22.0931 &            4.66805 & 2.500515\\
\hline
Li \cite{li2010simultaneous}: Dual Quaternion & 0.07 &   0.0000251894 &      0.185176 &          0.75443 &             15.9366 &    5.24936 & 14.9587\\
\hline
Li \cite{li2010simultaneous}: Kronecker product & 0.0001 & 0.0000227466 &            0.169176 &            0.972035 &             26.4559 &      5.67147 & 23.201702\\
\hline
Shah \cite{Shah2013Solving} & 0.0001 &  0.0000199547           & 0.149793             & 0.834843              & 19.515             &      4.6186 & 2.387767\\
\hline 
\end{tabular}
}
\end{center}
\label{table:dataset2}
\end{table*} 

\begin{table*}
\caption{Comparison of methods using the error metrics described in Section \ref{subsec:Experiments-ErrorMetrics} for Dataset 3.}
\begin{center}
\resizebox{\linewidth}{!}
{
\begin{tabular}{|c|S[table-format=3.4]|S[table-format=1.10]|S[table-format=2.6]|S[table-format=2.6]|S[table-format=6.3]|S[table-format=3.6]|S[table-format=6.5]|}
\hline
\textbf{Method} & \textbf{Time (s)} & \textbf{$e_{R1}$} & \textbf{$e_{R2}$} (degrees) &\textbf{$e_t$}(mm) & \textbf{$e_C$} & \textbf{$rrmse$}(pixels)& \textbf{$rae$}(mm)\\
\hline
Zhuang \cite{Zhuang1994simultaneous} & 0.013 & 0.765909     &          35.737 &                84.21 &               255288 &           1752.74 & \num{1.08832e+08} \\
\hline
Dornaika \cite{dornaika1998simultaneous}: Closed Form & 0.005 &  0.0760832 &              8.90491 &              97.4986 &               342216 &     1512.71 & \num{8.28684e+16}  \\
\hline
Dornaika \cite{dornaika1998simultaneous}: Iterative & 3.083 & 7.01367     &          139.069 &              75.6174 &               205854 &     1179.24 & \num{1.66168e+07}\\
\hline
Hirsh \cite{hirsh2001iterative} & 3.281 &  0.000146461       &       0.400659 &              6.17446 &              1372.46 &       39.071 & 3729.96\\
\hline
Li \cite{li2010simultaneous}: Dual Quaternion & 0.125 &  0.000229668    &           0.55199 &              5.34571 &              1028.76 &     20.7324 & 1896.01\\
\hline
Li \cite{li2010simultaneous}: Kronecker product & 0.002 & 00.000152553      &         0.42043 &              4.82104 &              836.727 &    31.7476 & 1832.47\\
\hline
Shah \cite{Shah2013Solving} & 0.0001 &  0.000146461    &          0.400666 &              2.76823 &              275.872 &       9.10238 & 123.061\\
\hline 
\end{tabular}
}
\end{center}
\label{table:dataset3}
\end{table*} 

\begin{table*}
\caption{Comparison of methods using the error metrics described in Section \ref{subsec:Experiments-ErrorMetrics} for Dataset 4.}
\begin{center}
\resizebox{\linewidth}{!}
{
\begin{tabular}{|c|S[table-format=3.4]|S[table-format=1.10]|S[table-format=2.6]|S[table-format=2.6]|S[table-format=6.3]|S[table-format=3.6]|S[table-format=6.6]|}
\hline
\textbf{Method} & \textbf{Time (s)} &\textbf{$e_{R1}$} & \textbf{$e_{R2}$} (degrees) &\textbf{$e_t$}(mm) & \textbf{$e_C$} & \textbf{$rrmse$}(pixels))& \textbf{$rae$}(mm)\\
\hline
Zhuang \cite{Zhuang1994simultaneous} &   0.003 &    0.00775597 &              2.41835               & 92.201 &               170021 &          288.166 & 36039.5\\ \hline
Dornaika \cite{dornaika1998simultaneous}: Closed Form & 0.003 &      0.010101 &              2.57979 &              122.624 & 300733 &         420.053 & 659311\\ \hline
Dornaika \cite{dornaika1998simultaneous}: Iterative &   1.692  &  7.97719 &              173.278 &              8.05142 &              1304.49 &              1235 & \num{1.11892e+17}\\ \hline
Hirsh \cite{hirsh2001iterative} &  3.409 &   0.0000954823 &             0.350316 &              6.06421 &              735.493 &          6.22611 & 76.2172\\ \hline
Li \cite{li2010simultaneous}: Dual Quaternion &   0.018 &   0.0675828       &        10.5063 &              63.6596 &              81050.9 &    189.191 &  232276\\ \hline
Li \cite{li2010simultaneous}: Kronecker product & 0.001 &    0.000120951    &          0.415269 &               7.2863 &               1061.8 &          21.69 & 4198.79\\ \hline
Shah \cite{Shah2013Solving} &   0.0001 &  0.0000954823 &             0.350316 &              3.12994 &              195.931 &    2.27361 & 22.7013\\
\hline
\end{tabular}
}
\end{center}
\label{table:dataset4}
\end{table*}  

\begin{table*}
\caption{Comparison of methods using the error metrics described in Section \ref{subsec:Experiments-ErrorMetrics} for Dataset 5.}
\begin{center}
\resizebox{\linewidth}{!}
{
\begin{tabular}{|c|S[table-format=3.4]|S[table-format=1.10]|S[table-format=2.6]|S[table-format=2.6]|S[table-format=6.3]|S[table-format=4.6]|S[table-format=4.6]|}
\hline
\textbf{Method}  & \textbf{Time (s)} &\textbf{$e_{R1}$} & \textbf{$e_{R2}$} (degrees) &\textbf{$e_t$}(mm) & \textbf{$e_C$} & \textbf{$rrmse$}(pixels) & \textbf{$rae$}(mm)\\
\hline
Zhuang \cite{Zhuang1994simultaneous} &    0.004 &  0.0118914   &            3.61164 &              113.639 &               180792          &       1190 & 862073\\ \hline
Dornaika \cite{dornaika1998simultaneous}: Closed Form & 0.002 &     0.0505229 &              7.58041 &              58.6203 &              48108.8 &    211.081 & 150077\\ \hline
Dornaika \cite{dornaika1998simultaneous}: Iterative &  1.116 &      4.36192   &            95.1924 &               31.289 &              13710.4 &    1206 & \num{1.36857e+10}\\ \hline
Hirsh \cite{hirsh2001iterative} &  0.354 &   0.0000261473  &            0.199554 &              152.636 &               326170 &      238.041 & \num{1.01182e+11}\\ \hline
Li \cite{li2010simultaneous}: Dual Quaternion &  0.008 &      0.0100287     &          4.05685 &              85.8321 &               103140 &   65.5218 & 62885.7\\ \hline
Li \cite{li2010simultaneous}: Kronecker product & 0.001 &      0.0000285929   &           0.203356 &              4.84945 &               329.24 &   7.16294 & 236.416\\ \hline
Shah \cite{Shah2013Solving} &  0.0001 &   0.0000261584     &     0.199265 &              210.092 &               617943 &          439.512 & \num{2.12246e+17}\\
\hline
\end{tabular}
}
\end{center}
\label{table:dataset5}
\end{table*}

\begin{table*}
\caption{Comparison of methods using the error metrics described in Section \ref{subsec:Experiments-ErrorMetrics} for Dataset 6. $rrmse$ is computed for each camera.}
\begin{center}
\resizebox{\linewidth}{!}
{
\begin{tabular}{|c|S[table-format=3.4] | S[table-format=1.10]|S[table-format=2.6]|S[table-format=2.6]|S[table-format=6.3]|S[table-format=1.6]|S[table-format=1.6]|S[table-format=6.6]|}
\hline
\textbf{Method} & \textbf{Time (s)} & \textbf{$e_{R1}$} & \parbox[t]{1cm}{\textbf{$e_{R2}$}\\(degrees)} &\textbf{$e_t$}(mm) & \textbf{$e_C$} & \parbox[t]{1cm}{\textbf{$rrmse_0$}\\(pixels)} & \parbox[t]{1cm}{\textbf{$rrmse_1$}\\(pixels)}& \textbf{$rae$}(mm)\\
\hline
\textbf{$c_1$, Euler angles, simultaneous} & 0.386 &  0.0000257326 &             0.193937 &              1.24552 &              21.0586 &                 2.31762 &              2.00706 &  16.6109\\ \hline
\textbf{$c_1$, axis angle, simultaneous} &  0.407 &   0.0000257326    &         0.193936 &              1.24552 &              21.0586 &            2.31762 &              2.00706 &  16.6109\\ \hline
\textbf{$c_1$, quaternion, simultaneous} &  0.424 &   0.0000257326  &           0.193936 &              1.24552 &              21.0586 &               2.31762 &              2.00706 &  16.6109\\ \hline
\textbf{$c_2$, Euler angles, simultaneous} & 0.242 & 0.0000280977 &             0.206962 &               1.9874 &              53.4269 &                     0.766459 &             0.774432 & 15.6238\\ \hline
\textbf{$c_2$, axis angle, simultaneous} & 0.248 &  0.0000249891   &           0.193788 &              1.90496 &              49.1943 &                   0.750998 &             0.734177 & 12.6656\\ \hline
\textbf{$c_2$, quaternion, simultaneous} & 0.262 & 0.000024989   &           0.193789 &              1.90496 &               49.194 &                    0.750993 &             0.734177 & 12.6686\\ \hline
\hline 
\textbf{$c_1$, Euler angles, separable} & 0.708 &  0.0000234965 &             0.186335 &               67.016 &              60837.1 &      68.7115 &              62.0413  &  141154 \\
\hline
\end{tabular}
}
\end{center}
\label{table:dataset6}
\end{table*}

\begin{table*}
\caption{Comparison of methods using the error metrics described in Section \ref{subsec:Experiments-ErrorMetrics} for Dataset 7. $rrmse$ is computed for each camera.}
\begin{center}
\resizebox{\linewidth}{!}
{
\begin{tabular}{|c|S[table-format=3.4]|S[table-format=1.10]|S[table-format=2.6]|S[table-format=2.6]|S[table-format=6.3]|S[table-format=1.6]|S[table-format=1.6]|S[table-format=1.6]|S[table-format=1.6]|}
\hline
\textbf{Method} & \textbf{Time (s)} & \textbf{$e_{R1}$} & \parbox[t]{1cm}{\textbf{$e_{R2}$}\\(degrees)} &\textbf{$e_t$}(mm) & \textbf{$e_C$} & \parbox[t]{1cm}{\textbf{$rrmse_0$}\\(pixels)} & \parbox[t]{1cm}{\textbf{$rrmse_1$}\\(pixels)} & \parbox[t]{1cm}{\textbf{$rrmse_2$}\\(pixels)} & \textbf{$rae$}(mm)\\
\hline
\textbf{$c_1$, Euler angles, simultaneous} &  1.548 &  0.000037808 &             0.223619 &              1.01891 &              37.8961 &     1.70065              & 1.70483            &   2.05211 & 25.5861\\ \hline
\textbf{$c_1$, axis angle, simultaneous} &  1.349 &     0.0000344302  &            0.216382 &              1.01891 &              37.8961 &                       1.76765 &              1.86726 &              1.57613 & 27.7002\\ \hline
\textbf{$c_1$, quaternion, simultaneous} &   1.13 &     0.00003705             & 0.219817              & 1.01891              & 37.8961                       & 1.59012              & 1.59645              & 1.89843 &  21.5528\\ \hline
\hline 
\textbf{$c_2$, Euler angles, simultaneous} & 1.273 & 0.0000380638             & 0.234888 &               1.3724 &              68.5225 &                         0.777057 &             0.689395 &             0.464227 & 9.86823\\ \hline
\textbf{$c_2$, axis angle, simultaneous} & 1.32 &  0.0000340799             & 0.214434              & 1.32509              & 63.8607                       & 0.740757             & 0.672692             & 0.464665 & 7.11799\\ \hline
\textbf{$c_2$, quaternion, simultaneous} & 1.656 & 0.0000340798      &        0.214439 &              1.32509 &              63.8605 &                         0.740759 &             0.672692 &             0.464664 & 7.12014\\
\hline
\end{tabular}
}
\end{center}
\label{table:dataset7}
\end{table*}

\begin{table*}
\caption{Comparison of methods using the error metrics described in Section \ref{subsec:Experiments-ErrorMetrics} for Dataset 8. $rrmse$ is computed for each camera.}
\begin{center}
\resizebox{\linewidth}{!}
{
\begin{tabular}{|c|S[table-format=3.4]|S[table-format=1.10]|S[table-format=2.6]|S[table-format=2.6]|S[table-format=6.3]|S[table-format=1.6]|S[table-format=1.6]|S[table-format=1.6]|S[table-format=1.6]|}
\hline
\textbf{Method} & \textbf{Time (s)} & \textbf{$e_{R1}$} & \parbox[t]{1cm}{\textbf{$e_{R2}$}\\(degrees)} &\textbf{$e_t$}(mm) & \textbf{$e_C$} & \parbox[t]{1cm}{\textbf{$rrmse_0$}\\(pixels)} & \parbox[t]{1cm}{\textbf{$rrmse_1$}\\(pixels)} & \parbox[t]{1cm}{\textbf{$rrmse_2$}\\(pixels)} & \textbf{$rae$}(mm)\\
\hline

\textbf{$c_1$, Euler angles, simultaneous} & 1.372 & 0.0000332003 &             0.208999 &              1.03046 &              34.7303 &     2.51556 &              3.04886 &              1.38591 & 35.3555\\ \hline
\textbf{$c_1$, axis angle, simultaneous} & 1.191 &  0.0000215955    &          0.172318 &              1.03046 &              34.7303 &   1.47545 &              1.74632 &             0.830485 & 9.76088\\ \hline
\textbf{$c_1$, quaternion, simultaneous} & 0.998 & 0.0000247851    &          0.177514 &              1.03046 &              34.7303 &       1.67143 &              1.81211 &             0.984151 & 10.5367\\ \hline
\hline 
\textbf{$c_2$, Euler angles, simultaneous} & 1.114 & 0.0000330719 &             0.221522 &              1.42353 &              65.1598 &    0.805084 &             0.885025 &             0.429181 & 3.74888\\ \hline
\textbf{$c_2$, axis angle, simultaneous} &  1.164 &    0.0000288111  &            0.197555 &              1.36356 &              60.1969 &   0.776654 &             0.861649 &             0.422205 & 2.18689\\ \hline
\textbf{$c_2$, quaternion, simultaneous} &  1.452 &  0.0000288111    &          0.197561 &              1.36355 &              60.1956 &  0.776668 &             0.861652 &             0.422208 & 2.18551\\
\hline
\end{tabular}
}
\end{center}
\label{table:dataset8}
\end{table*}

\section{Conclusions} \label{sec:conclusions}

We presented a collection of methods for robot-world hand-eye calibration (including the case of multiple cameras) with accompanying code.  Within this collection, we explored cost functions related to the equality $\mathbf{AX}= \mathbf{ZB}$ (including simultaneous and separable versions) and camera reprojection error.  We also implemented these minimization problems using three different rotation representations.  Our collection of methods was evaluated relative to existing robot-world hand-eye calibration methods.

This collection of methods was extended to be used to calibrate multiple cameras mounted on one robot, and this was demonstrated on datasets with one to three cameras. To the best of our knowledge, there are currently no other robot-world hand-eye calibration methods for multiple cameras, and consequently we believe this work is the first attempt to solve such a problem. We also introduced a metric, called Reconstruction Accuracy Error, or $rae$, to assess a particular method's suitability to reconstruction problems.  

Our collections of methods, particularly the $c_2$ simultaneous method from the first class and both methods in the second class perform consistently well with respect to reconstruction accuracy error, generating $rae \leq 2$mm, which is the average distance between a calibration pattern point and the point generated with the robot-world, hand-eye calibration information.  This value of $rae \leq 2$mm resulted from evaluating the collection of methods on a range of real datasets representing different mounting configurations and robot positions relative to the world coordinate frame.

\begin{acknowledgements}
We also would like to thank anonymous reviewers of our previous paper \cite{Tabb2015Parameterizations} and this paper for their helpful comments.
\end{acknowledgements}

\bibliographystyle{spmpsci}
\bibliography{sources_hand_eye}

\begin{thebibliography}{10}
\providecommand{\url}[1]{{#1}}
\providecommand{\urlprefix}{URL }
\expandafter\ifx\csname urlstyle\endcsname\relax
  \providecommand{\doi}[1]{DOI~\discretionary{}{}{}#1}\else
  \providecommand{\doi}{DOI~\discretionary{}{}{}\begingroup
  \urlstyle{rm}\Url}\fi

\bibitem{opencv_library}
Opencv.
\newblock \verb+http://opencv.org/+.
\newblock Version 2.4.9

\bibitem{ceressolver}
Agarwal, S., Mierle, K., Others: Ceres solver.
\newblock \url{http://ceres-solver.org}

\bibitem{dornaika1998simultaneous}
Dornaika, F., Horaud, R.: Simultaneous robot-world and hand-eye calibration.
\newblock Robotics and Automation, IEEE Transactions on \textbf{14}(4),
  617--622 (1998)

\bibitem{hirsh2001iterative}
Hirsh, R.L., DeSouza, G.N., Kak, A.C.: An iterative approach to the hand-eye
  and base-world calibration problem.
\newblock In: Robotics and Automation, 2001. Proceedings 2001 ICRA. IEEE
  International Conference on, vol.~3, pp. 2171--2176. IEEE (2001)

\bibitem{Horaud1995Hand}
Horaud, R., Dornaika, F.: Hand-eye calibration.
\newblock The international journal of robotics research \textbf{14}(3),
  195--210 (1995)

\bibitem{li2010simultaneous}
Li, A., Lin, W., Defeng, W.: Simultaneous robot-world and hand-eye calibration
  using dual-quaternions and kronecker product.
\newblock International Journal of Physical Sciences \textbf{5}(10), 1530--1536
  (2010)

\bibitem{lourakis04LM}
Lourakis, M.: levmar: Levenberg-marquardt nonlinear least squares algorithms in
  {C}/{C}++.
\newblock [web page] \verb+http://www.ics.forth.gr/~lourakis/levmar/+ (Jul.
  2004).
\newblock [Accessed on 18 Nov. 2014.]

\bibitem{malti2013hand}
Malti, A.: Hand--eye calibration with epipolar constraints: Application to
  endoscopy.
\newblock Robotics and Autonomous Systems \textbf{61}(2), 161--169 (2013)

\bibitem{marquardt1963algorithm}
Marquardt, D.W.: An algorithm for least-squares estimation of nonlinear
  parameters.
\newblock Journal of the Society for Industrial \& Applied Mathematics
  \textbf{11}(2), 431--441 (1963)

\bibitem{Shah2013Solving}
Shah, M.: Solving the robot-world/hand-eye calibration problem using the
  kronecker product.
\newblock ASME Journal of Mechanisms and Robotics \textbf{5}(3),
  031,007--031,007--7 (2013)

\bibitem{shiu1989calibration}
Shiu, Y.C., Ahmad, S.: Calibration of wrist-mounted robotic sensors by solving
  homogeneous transform equations of the form ax= xb.
\newblock Robotics and Automation, IEEE Transactions on \textbf{5}(1), 16--29
  (1989)

\bibitem{Strobl2006Optimal}
Strobl, K.H., Hirzinger, G.: Optimal hand-eye calibration.
\newblock In: 2006 IEEE/RSJ International Conference on Intelligent Robots and
  Systems, pp. 4647--4653 (2006).
\newblock \doi{10.1109/IROS.2006.282250}

\bibitem{Tabb2013}
Tabb, A.: Shape from silhouette probability maps: reconstruction of thin
  objects in the presence of silhouette extraction and calibration error.
\newblock In: Computer Vision and Pattern Recognition (CVPR), 2013 IEEE
  Conference on (2013)

\bibitem{tabb2017solving_dataset}
Tabb, A.: Data from: Solving the robot-world hand-eye(s) calibration problem
  with iterative methods (2017).
\newblock \doi{10.15482/USDA.ADC/1340592}.
\newblock \urlprefix\url{http://dx.doi.org/10.15482/USDA.ADC/1340592}

\bibitem{Tabb2015Parameterizations}
Tabb, A., Ahmad~Yousef, K.: Parameterizations for reducing camera reprojection
  error for robot-world hand-eye calibration.
\newblock In: IEEE RSJ International Conference on Intelligent Robots and
  Systems, pp. 3030--3037 (2015)

\bibitem{Triggs1999Bundle}
Triggs, B., McLauchlan, P.F., Hartley, R.I., Fitzgibbon, A.W.: Bundle
  adjustment - a modern synthesis.
\newblock In: Proceedings of the International Workshop on Vision Algorithms:
  Theory and Practice, ICCV '99, pp. 298--372. Springer-Verlag, London, UK, UK
  (2000).
\newblock \urlprefix\url{http://dl.acm.org/citation.cfm?id=646271.685629}

\bibitem{zhang2000flexible}
Zhang, Z.: A flexible new technique for camera calibration.
\newblock Pattern Analysis and Machine Intelligence, IEEE Transactions on
  \textbf{22}(11), 1330--1334 (2000).
\newblock \doi{10.1109/34.888718}

\bibitem{Zhuang1994simultaneous}
Zhuang, H., Roth, Z.S., Sudhakar, R.: Simultaneous robot/world and tool/flange
  calibration by solving homogeneous transformation equations of the form
  ax=yb.
\newblock Robotics and Automation, IEEE Transactions on \textbf{10}(4),
  549--554 (1994).
\newblock \doi{10.1109/70.313105}

\end{thebibliography}

\clearpage

\smallbreak

\twocolumn[\section{Erratum} \label{changelog}

We regret that there were some errors in the original paper, concerned mostly with the computation of the mean translation error $e_t$ (Section \ref{ss:mean_trans}, Equation \ref{eq:transError}).  

Specifically, 
\begin{enumerate}
\item{The computation of $e_t$ was incorrect in our software used to generate the results in tables \ref{table:dataset1}-\ref{table:dataset8} in the original paper.}
\item{The labeling of units of $e_t$ was incorrect; it should have been mm$^2$.}
\item{Tables with corrected results are included in this erratum section, in tables \ref{table:dataset1_erratum}-\ref{table:dataset8_erratum}.}
\item{The software was released with the paper at \cite{tabb2017solving_dataset}; this software has been upgraded to present-day libraries and the error in computing $e_t$ noted here has been corrected.}
\item{Since the underlying libraries supporting the software has undergone revision since we prepared the results for the original paper, there are slight differences in the numerical results despite using the same input.  Consequently, we included completely refreshed results in tables \ref{table:dataset1_erratum}-\ref{table:dataset8_erratum}.}
\end{enumerate}]

\begin{table*}
\caption{Erratum: Comparison of methods using the error metrics described in Section \ref{subsec:Experiments-ErrorMetrics} for Dataset 1.}
\begin{center}
\resizebox{\linewidth}{!}
{
\begin{tabular}{|c|S[table-format=3.3]|S[table-format=1.6]|S[table-format=2.6]|S[table-format=6.3]|S[table-format=6.3]|S[table-format=3.6]|S[table-format=8.6]|}
\hline
\textbf{Method} & \textbf{Time (s)} & \textbf{$e_{R1}$} & \textbf{$e_{R2}$}(degrees) &\textbf{$e_t$}(mm$^2$) & \textbf{$e_C$} & \textbf{$rrmse$}(pixels) & \textbf{$rac$}(mm)\\
\hline
Zhuang \cite{Zhuang1994simultaneous} &   0.0210    &         0.000101     &          0.33641     &          233.243     &          233.243     &          4.35291 &                         434.01527\\
\hline
Dornaika \cite{dornaika1998simultaneous}: Closed Form &   0.0050    &         0.261051     &         19.51617     &       330046.750     &       330047.011     &        761.54201 &                      158329.92262\\
\hline
Dornaika \cite{dornaika1998simultaneous}: Iterative &   3.0130     &         1.947004     &         58.65204     &       283023.041     &       283024.988     &        272.69726 &    \num{2.3137e+08}\\
\hline
Hirsh \cite{hirsh2001iterative} &   1.847    &         0.000098     &          0.33164     &          359.328     &          359.328     &          3.68832 &                         189.53753\\
\hline
Li \cite{li2010simultaneous}: Dual Quaternion &   0.479    &         0.000174     &          0.50621     &          320.173     &          320.174     &          4.30732 &                         103.75765\\
\hline
Li \cite{li2010simultaneous}: Kronecker product &   0.002    &         0.000102     &          0.34782     &          446.079     &          446.080     &          5.41762 &                         388.56442\\
\hline
Shah \cite{Shah2013Solving} &   0.002     &         0.000098     &          0.33164     &          235.577     &          235.577     &          4.10188 &                         403.65687\\
\hline
\end{tabular}
}
\end{center}
\label{table:dataset1_erratum}
\end{table*}
  
\begin{table*}
\caption{Erratum: Comparison of methods using the error metrics described in Section \ref{subsec:Experiments-ErrorMetrics}  for Dataset 2.}
\begin{center}
\resizebox{\linewidth}{!}
{
}
\end{center}
\label{table:dataset2_erratum}
\end{table*} 

\begin{table*}
\caption{Erratum: Comparison of methods using the error metrics described in Section \ref{subsec:Experiments-ErrorMetrics}  for Dataset 3.}
\begin{center}
\resizebox{\linewidth}{!}
{
}
\end{center}
\label{table:dataset3_erratum}
\end{table*}

\begin{table*}
\caption{Erratum: Comparison of methods using the error metrics described in Section \ref{subsec:Experiments-ErrorMetrics}  for Dataset 4.}
\begin{center}
\resizebox{\linewidth}{!}
{
}
\end{center}
\label{table:dataset4_erratum}
\end{table*}

\begin{table*}
\caption{Erratum: Comparison of methods using the error metrics described in Section \ref{subsec:Experiments-ErrorMetrics}  for Dataset 5.}
\begin{center}
\resizebox{\linewidth}{!}
{
}
\end{center}
\label{table:dataset5_erratum}
\end{table*}

\begin{table*}
\caption{Erratum: Comparison of methods using the error metrics described in Section \ref{subsec:Experiments-ErrorMetrics}  for Dataset 6. $rrmse$ is computed for each camera}
\begin{center}
\resizebox{\linewidth}{!}
{
}
\end{center}
\label{table:dataset6_erratum}
\end{table*}

\begin{table*}
\caption{Erratum: Comparison of methods using the error metrics described in Section \ref{subsec:Experiments-ErrorMetrics}  for Dataset 7. $rrmse$ is computed for each camera}
\begin{center}
\resizebox{\linewidth}{!}
{
}
\end{center}
\label{table:dataset7_erratum}
\end{table*}

\begin{table*}
\caption{Erratum:Comparison of methods using the error metrics described in Section \ref{subsec:Experiments-ErrorMetrics}  for Dataset 8. $rrmse$ is computed for each camera}
\begin{center}
\resizebox{\linewidth}{!}
{
}
\end{center}
\label{table:dataset8_erratum}
\end{table*}

\end{document}